\documentclass[12pt]{article}
\usepackage[top=1in, bottom=1in, left=1in, right=1in]{geometry}
\usepackage{geometry}
\usepackage{amsthm, soul, amsfonts}
\usepackage{amsmath, amssymb}
\usepackage{authblk, array, setspace} 
\usepackage{natbib}
\usepackage[pdftex,colorlinks=true,linkcolor=blue,citecolor=blue,urlcolor=blue,bookmarks=false,pdfpagemode=None]{hyperref}
\usepackage{enumerate}
\usepackage{graphicx}
\usepackage{float}
\usepackage{pgffor}
\newtheorem{theorem}{Theorem}[section]
\newtheorem{lemma}[theorem]{Lemma}

\usepackage{hyperref}
\usepackage{algorithm}
\usepackage{algorithmic}

\title{
Multiple Metric Learning for Structured Data
}
\author{nicolo colombo}

\affil{Department of Computer Science, 
Royal Holloway University of London, 
Egham, United Kingdom \footnote{nicolo.colombo@rhul.ac.uk}}

\begin{document}

\maketitle
\begin{abstract}

	We address the problem of merging 
	graph and feature-space information while learning 
	a metric from structured data. 	
	Existing algorithms tackle the problem
	in an asymmetric way, by either extracting vectorized 
	summaries of the graph structure or adding  
	hard constraints to feature-space algorithms.\\
	Following a different path, we
	define a metric regression scheme where we train 
	metric-constrained linear combinations
	of dissimilarity matrices.
	The idea is that the input matrices can be pre-computed 
	dissimilarity measures obtained  
	from any kind of available data (e.g. node attributes 
	or edge structure). 
	As the model inputs are distance measures, 
	we do not need to assume the 
	existence of any underlying feature space.
	Main challenge is that metric constraints (especially positive 
	definiteness and sub-additivity), are not 
	automatically respected if, for example, 
	the coefficients of the linear combination 
	are allowed to be negative.
	Both positive and sub-additive constraints 
	are linear inequalities, 
	but the computational complexity 
	of imposing them scales as $O(D^3)$, 
	where $D$ is the size of the input matrices  
	(i.e. the size of the data set). 
	This becomes quickly prohibitive, even when
	$D$ is relatively small.
	We propose a new graph-based technique for optimizing 
	under such constraints and show that, 
	in some cases, our approach may 
	reduce the original computational complexity of the 
	optimization process by one order of magnitude.
	Contrarily to existing methods, our scheme 
	applies to any (possibly non-convex)  
	metric-constrained objective function. 


\end{abstract}
\section{Introduction}
\subsection{Motivation}
\label{section motivations}
In many applications, e.g. social networks analysis,
data sets are structured objects consisting of 
text, real or Boolean high dimensional vectors and a graph.
Text and vectors usually provide a description of the real-world 
entities represented by the graph nodes, 
while the edge structure  
is associated with some kind of relationship 
between them.
Through standard word-embedding techniques, 
it is possible, and also pretty efficient, to transform 
node-related texts to vectors.
Word-embedding vectors can then be  
merged with real-valued node attributes by 
opportunely enlarging the feature space.
Taking into account the graph edge structure 
is in general less straightforward.
This because, in most cases, the 
complex relationships represented by graph links 
cannot be reduced to simple Euclidean 
distances defined over an explicit feature space. 

Goal of this work is to present a new and general method for 
learning statistical models from graph-structured data sets.
We exploit some crucial properties of 
metric spaces to define an optimization scheme 
that encodes the whole knowledge provided by a data set into 
a single mathematical object.
Obtained from both node and edge features, the latter can be 
seen as the metric of the entire data set.
The obtained data-set metric can be used directly to solve classical 
machine learning tasks
through extremely simple algorithms such as $(k=1)$ Nearest Neighbor 
or K-Means algorithms. 

The inspiring general idea is that real-world systems
(e.g. social networks, financial markets or power energy grids)
can be represented in a geometric way by 
abstract (and not necessarily Euclidean) manifolds.
Data points obtained from the systems 
sit on such abstract manifolds and can be used to  
detect their underlying (metric) structure.
In the machine learning literature, 
the task of inferring metric structures from data 
is referred to as the metric learning problem.
In the same context, multiple metric learning 
is a promising and recent extension of the classical 
metric learning setup where 
the task is to optimize linear combinations of 
different distance measures.

Existing approaches mostly focus on Mahalanobis metrics. 
In that case, data points are mapped to feature vectors 
lying on a high dimensional space.
As the high-dimensional space is usually equipped
with a flat Euclidean metric, 
metric learning reduces to inferring an optimal 
linear transformation between the original data points   
and such high dimensional vectors. 
Advantages of the Mahalanobis approach are 
the convexity of the optimization problem and 
the fact that the output is automatically 
guaranteed to be a metric, 
e.g. it is positive semi-definite and fulfils all triangle 
inequalities $M_{ij} \leq M_{ik} + M_{kj}$ ($i,j,k=1, \dots, D$).
Technically, this happens because the feature-space metric 
is defined by $M_{ij} = \| x^{(i)} - x^{(j)} \|^2$($i, j=1, \dots, D$), 
where 
$x^{(i)}$  are the high-dimensional 
representatives of the data points  
and $\| x \|^2 = x^T x$ is the Euclidean norm, which satisfies the 
triangle inequality by definition.

As mentioned above, however, methods based on Euclidean distances 
struggle when the input data set contains structured information.
The reason is that graph edges are not 
always defined as Euclidean distances between explicit 
feature-space vectors, i.e. it does not always 
exist $\{ z^{(i)} \in {\mathbf R}^{N}\}_{i=1}^D$ ($N \in {\mathbf N}_+$) 
such that $M_{ij} = \| z^{(i)} - z^{(j)}\|^2$ 
for all (given) edge weights $M_{ij}$ ($i,j=1, \dots, D$) .
We tackle this problem by proposing a 
two-step scheme where we first extract a set of metrics 
from the structured data set (e.g. one distinct metric 
for each different piece of information in the data set)
and then consistently combine them into an optimal mixture
by solving a metric-constrained optimization process.

\subsection{Related work}
\label{section related work}
\paragraph{Metric constraints as a regularization scheme}
Metric constraints can be looked at as 
an interpretable regularization scheme to reduce the complexity 
of a statistical model.
Practical implementations of this idea have already 
appeared in different domains.
It is widely known, for example, that imposing 
metric constraints in k-means clustering can be highly 
effective in terms of computational 
complexity and accuracy of the output 
\citep{xing2003distance, bilenko2004integrating, weinberger2009distance}. 
More broadly, dissimilarity measures and 
metrics have been shown to be important 
for shape analysis \citep{basri1998determining}, object
recognition \citep{bronstein2007rock} and features 
selection \citep{laub2004feature}. 
\cite{memoli2011metric} analyzes the 
role of metrics in data science and 
discusses their interpretation as summaries of complex datasets. 
\paragraph{Metric learning and structured data}
The task of unveiling the metric structure of
a given set of observations is known in the literature
as metric learning \citep{kulis2013metric} and has many practical 
applications (see \cite{yu2006toward} or
\cite{hoi2006learning} for an example in computer vision).
The connection with graphs and structured data have
also been investigated in the past.
\cite{hauberg2012geometric}
focuses on computing geodesic on a feature space where 
the metric is defined as the superposition of metrics.
This is similar to  
our approach
but the formulation of \cite{hauberg2012geometric}
only applies to continuous feature spaces.
The paper also shows that being a non-metric may
have effects on certain applications.
\cite{shi2014sparse}
is a work very close to ours.  
The authors explain how to learn combinations 
of global and local metrics.
They apply the scheme to multi-task learning problems, which  
may be solved in a very similar way through the approach proposed here.
There are, however, few important differences: 
i) the paper focuses on Mahalanobis metrics, and hence assume the 
existence of an Euclidean feature space;  
ii) the optimized linear combinations have only non-negative weights, 
which means that the output is guaranteed to be a metric. 
As in our experiments, the authors evaluate their method through 
a simple Nearest Neighbor algorithm.
\cite{shaw2011learning} is an example of how
 the graph structure can be enforced imposing by hard constraints. 
 The learning task considered in \cite{shaw2011learning} is 
 slightly different (the 
 authors learn a metric 
 ``such that points connected in the network are close and 
points which are unconnected are more distant.'')
but the same approach may be used in different setups
without major modifications.
\cite{shaw2009structure}
presents one of the several existing algorithms for 
embedding graphs in Euclidean spaces.
In that work, the embedding is guaranteed to be 
low dimensional and to preserve the global 
topological properties of the input graph.
Our approach does not need any embedding step 
but comparing our method to those approaches
would be interesting.\footnote{As it can be seen 
from the simple experiments 
described in Section \ref{section experiments}, 
the main purpose of this paper is 
methodological. 
We leave an extensive comparison
of our approach with the state-of-the-art for the 
next future.
}
\cite{bellet2013survey}
is a good survey of metric learning methods that assume the existence 
of an underlying feature space.
Many of these rely on converting structured objects 
to feature vectors through various techniques 
(as for example the string kernel method 
of \cite{lodhi2002text} or the bag-of-words approach of
\cite{salton1975vector} and \cite{fei2005bayesian}).
Section 5 of \cite{bellet2013survey} contains an
exhaustive list of metric learning algorithms that 
can handle structured data without an explicit vector 
representation.
Most of these algorithms, however, are designed specifically for 
text-based data and exploit some very peculiar features
of the `edit distance'.

\paragraph{Metric-constrained optimization} Metric learning is closely related 
but not equivalent to metric-constrained optimization. 
As mentioned above metric learning is the problem of inferring a 
distance measure from a given set of observations 
\citep{xing2003distance, kulis2013metric}.
In the classical setup of metric learning, the starting point 
is an accessible feature space (from which the target metric is built).
Metric-constrained optimization problems includes the more 
general situation where the feature space can be implicit or not accessible.
A more similar problem is multidimensional scaling, where the task is
to find N low-dimensional points such that their distances respect 
a set of given dissimilarity relationships \citep{cox2000multidimensional}.
In a typical metric-constrained optimization problem, however, 
the variable is another implicit distance measure and not 
an explicit low-dimensional features (as in multidimensional scaling).
In \cite{veldt2018projection} is one of the few example of
work about metric-constrained optimization. 
The authors propose a projection method based 
on the Dikjstra algorithm \citep{dykstra1983algorithm} and 
test it on various tasks, such as correlation clustering 
\citep{batra2008semi}, sparsest or maximum cut 
and the metric nearness problem (see below).
All these applications can be reformulated as a linear or quadratic 
programming with $O(N^3)$ triangle-inequality linear constraints.\footnote{
	Note that the Dikjstra algorithm, which the projection 
method of \cite{veldt2018projection} is based on, 
can be only used in least-squares or linear problems.}
\cite{roth2003going} is another work that explicitly addresses 
a metric-constrained optimization problem.
In that case, metric constraints are enforced by learning 
a global transformation of a given set of pairwise distances.  
Optimization under triangle-inequality 
constraints have also appeared in various management applications 
\citep{igelmund1983preselective, hanson1990optimal}, but mostly for 
integer programming \citep{burdet1975subadditive}.

\paragraph{Intrinsic metrics}
The use of intrinsic and extrinsic features of a given 
data set has been used for improving the performance of 
various algorithms.
\cite{bronstein2007rock} contains an interesting comparison between 
intrinsic and extrinsic similarities, in the framework of shape analysis.
A more contextualized description of intrinsic metrics 
in data science can be found in \cite{keller2015intrinsic}.
Here we refer to a special case of discrete intrinsic metric, also known as 
path-length distance.
See for example \cite{buckley1990distance, diestel2018graph, 
west2001introduction} or \cite{bandelt2008metric} for a general survey on graph metrics.
A crucial feature of path-length distances 
is that they automatically respect all triangle inequalities. 
Actually, they are the discrete versions of intrinsic metrics 
defined over continuous spaces, which  
are often used to characterize a manifold without 
choosing a particular embedding space (see for \cite{burago2001course} or \cite{bridson2013metric} for more details).

\paragraph{Multiple kernel learning}
Multiple kernel learning (see for example 
\cite{lanckriet2004learning}) has some strong analogy with 
our approach. 
What makes metric learning more challenging 
is the presence of the $O(D^3)$ triangle-inequality 
constraints.
As for metrics, 
the positive definiteness constraint of 
multiple kernel learning is automatically 
satisfied when the output is a linear combination 
with non-negative weights.\footnote{This is not true 
if the non-negative linear combination 
includes dissimilarity matrices (for metrics) or
symmetric positive-undefined matrices (for kernels).}
In our work, we focus for simplicity on linear 
combinations of metrics but drop the non-negativity 
restriction to show the potential generality of the approach. 

\paragraph{Metric nearness problem}
The task of finding
the closest metric to a given dissimilarity matrix 
has been called the metric nearness problem.
A triangle-fixing algorithm for solving this specific problem 
is proposed and analyzed theoretically in \cite{brickell2008metric}.
Based on this, \cite{veldt2018projection}
introduces a more general metric-projection algorithm to solve 
linear and quadratic optimization problems.

\section{Methods}

\paragraph{Notation}
We use capital letters with lowercase indexes 
for matrices, e.g. $X_{ij} = [X]_{ij}$ 
($i,j=1, \dots, D$).
Lowercase letters with capital indexes are used for their 
vectorized form, i.e. $x_I = [{\rm vec}(X)]_I$ ($I=1, \dots, D^2$) and
$X_{ij} = x_I$ for $I = D(i-1) + j$.
Element wise product is denoted by ``$\circ$'' and the transpose of  
$x$ by $x^T$.
${\cal M}_D \subset {\mathbf R}^{D \times D}$ is the space 
of $D \times D$ metrics, i.e. the space of real matrices that obey 
all metric constraints \eqref{metric properties}.
The following short notation is used for linear combinations of matrices 
$M_\alpha = \sum_{r=1}^R \alpha_r M_r \in {\mathbf R}^{D \times D}$ and 
vectors
$m_\alpha = \sum_{r = 1}^R \alpha_r m_r \in {\mathbf R}^{D^2}$.
Given a symmetric matrix, $A = A^T$, ${\cal G}(A)$ is 
the undirected graph obtained by interpreting $A$ as 
the graph adjacency matrix.
$\Gamma_{ij} = \Gamma_{ij}({\cal G})$ is the set of simple paths 
connecting nodes $i$ and $j$ on ${\cal G}$.  
Let $p$ be a path on ${\cal G}$, then 
$\gamma = \gamma(p) \in \{ 0, 1 \}^{D^2}$ 
is the vector (unordered) 
representation of $p$ defined by 
\begin{eqnarray}
	\left[ \gamma(p) \right]_I = \left\{ 
	\begin{array}{ll}
		1 & {\rm if \ }  {\rm link}_{i\to j} 
		\in p {\rm \ and  \ } I = D(i-1) + j \\
		0 & {\rm otherwise}
	\end{array}
	\right. ,\nonumber  
\end{eqnarray} 
where ${\rm link}_{i\to j}$ is the link between nodes $i$ and $j$ on ${\cal G}$.
As a consequence, the length of $p$ on ${\cal G}(A)$ is $\gamma^T m$.
More generally, the path-distance between nodes $i$ and $j$ 
on graph ${\cal G}$ is defined by 
${\rm dist}_{\rm path}(i, j, {\cal G}) = 
\min_{{\rm path} \in \Gamma_{ij}({\cal G})} {\rm length}({\rm path})$.
$\nabla f$ is th he gradient of $f:{\mathbf R}^N \to {\mathbf R}$ defined 
by $[\nabla f]_{r} = \frac{\partial f(x)}{\partial x_r}$.
The `softplus' function and its derivatives are  
${\rm softplus}(x) = \log(1 + e^{x})$,  
$\partial_x {\rm softplus}(x) = \sigma(x) = (1 + e^{-x})^{-1}$ and 
$\partial_x \sigma(x) = \sigma'(x)$ and 
apply element wise to matrices and vectors.

\subsection{Metric spaces}
A discrete metric space is a discrete set,  
${\cal D} = \{ x^{(i)} \}_{i=1}^D$, and a metric,  $M \in {\cal M}$, 
defined on ${\cal D}$.
$M$ is a metric on ${\cal D}$ if
\footnote{
	Pseudo metrics and non-symmetric metrics have been also considered, 
	see for example \cite{zaustinsky1959spaces, mennucci2004asymmetric}.
}
\begin{eqnarray}
	\label{metric properties}
	M_{ij} &=& M_{ji}, \\
	M_{ij} &>& 0 {\rm \ if} \ i \neq j, \nonumber \\ 
	M_{ii} &=& 0 \nonumber \\
	M_{ij} &\leq& M_{ik} + M_{kj},  \nonumber  
\end{eqnarray}
for all $i,j, k=1, \dots, D$.
The last inequality is referred to as triangle inequality. 
$M \in {\cal M}$ can be seen as 
the adjacency matrix of a fully-connected, 
undirected and weighted graph, ${\cal G}(M)$. 
Conversely, given a connected (but not necessarily fully-connected)
graph, ${\cal G}$, 
it is always possible to define a metric over its node set  
by letting 
\begin{eqnarray}
\label{path length metric}
	\tilde M_{ij} = {\rm dist}_{\rm path}(i, j, {\cal G}),   
\end{eqnarray}
for all $i,j =1, \dots, D$.
In this case, $\tilde M$ is the intrinsic metric 
of ${\cal G}$.
It is easy to verify that $\tilde M \in {\cal M}$.
\footnote{The intrinsic metric of a graph  
does not always coincide with its adjacency matrix.
}
A metric space, $({\cal D}, M)$ is 
a length metric space if $M = \tilde M$.
For example, ${\mathbf R}^N$ equipped with the Euclidean  
metric, which is defined by $M_{ij} = \|x^{(i)} - x^{(j)} \|^2$ 
($x^{(i)} \in {\mathbf R}^N$, $i,j=1, \dots D$), is a length metric space.

\subsection{Metric-constrained optimization}
\label{section metric constrained optimization}
We address the problem of optimizing a scalar function, 
$f(M) : {\mathbf R}^{D \times D} \to {\mathbf R}$,   
under the constraint $M \in {\cal M}$.
This can be performed explicitly, by imposing the 
$O(n^3)$ linear constraints \eqref{metric properties} during  
the optimization (e.g. through a projection method), 
or implicitly, by constraining the matrix 
variable to be a path-distance (intrinsic) metric.
\footnote{A shortest-path approach is also mentioned 
in \cite{brickell2008metric}. 
The authors show that finding the closest intrinsic 
metric to a given dissimilarity 
matrix, $X$, is equivalent to
solving a metric nearness problem
under the further 
constraint $X_{ij} > M_{ij}$, 
which is referred to as the decrease only metric nearness problem. 
The equivalence between has 
analyzed further in \cite{williams2018subcubic}, which 
focuses on a computational complexity perspective.
In \cite{brickell2008metric}, the result is used  
to solve the all-pairs shortest path problem 
through linear programming algorithms.
As mendtioned in Section \ref{section related work}, the projection scheme 
of \cite{brickell2008metric}, that is based on the Dijkstra
algorithm, requires the objective function to be linear (or at most quadratic).
}

\paragraph{Direct methods}
When $f$ is a general function, 
metric-constrained problems as $\min_{M \in {\cal M}} f(M)$ 
can be solved by usual algorithms designed for    
inequality-constrained optimization by imposing 
\eqref{metric properties}
directly (e.g. through a barrier method).
Enforcing \eqref{metric properties}, however, 
scales as $O(D^3)$ and becomes unfeasible even 
for relatively small $D$.
During the numerical simulations described in 
Section \ref{section experiments},  for example, 
we have run (on a standard laptop) a Scipy package 
for constrained optimization 
and incurred in memory errors for $D > 70$.

\paragraph{Intrinsic-metric projection method}
To tackle the scalability problem mentioned above, 
we focus on functions of the form 
\begin{equation}
	\label{separable function}
	f(M) = \sum_{i,j=1}^D g(M_{ij}), \quad g:{\mathbf R} \to {\mathbf R}, 
\end{equation} 
Our optimization scheme can be summarized as follows.

{\bf Step 1} Consider a shortest-path projector
\footnote{Equivalently, an intrinsic-metric projector.
},  
${\cal P}$, defined by
\begin{equation}
	\label{projector}
	[{\cal P}(X)]_{ij} = \min_{\gamma \in \Gamma_{ij}} 
	\gamma^T {\rm softplus}(x), 
	\quad \Gamma_{ij} = \Gamma_{ij}({\rm softplus}(X))
\end{equation}
where $i,j=1, \dots, D$ and $X \in {\mathbf R}^{D \times D}$ obeys 
is $X^T = X$.

{\bf Step 2} Transform the original metric-constrained problem 
into an unconstrained problem
\footnote{The symmetry constraint, $X = X^T$, is easy to be taken 
into account. 
For example, we can define
\begin{equation}
	[{\cal P}(X)]_{ij} = \min_{\gamma \in \Gamma_{ij}} 
	\gamma^T {\rm softplus}({\rm vec}(\frac{X + X^T}{2})), 
	\quad \Gamma_{ij} = \Gamma_{ij}({\rm softplus}(\frac{X + X^T}{2})), 
\end{equation}
where $i,j=1, \dots, D$ and $X \in {\mathbf R}^{D \times D}$, 
and then drop the symmetry constraint in \eqref{unconstrained problem}.
In all experiments presented in Section \ref{section experiments}, we 
never need to impose $X = X^T$ explicitly as all input matrices 
are metrics or dissimilarity measures, which are 
symmetric by definition. 
Finally, we note that the intrinsic projector \eqref{projector} 
applies without changes to non-symmetric metrics 
\citep{zaustinsky1959spaces, mennucci2004asymmetric}. 
It would be interesting 
to test the proposed method on related applications, e.g. road planning 
on complex transportation networks.}
\begin{eqnarray}
	\label{unconstrained problem}
	\min_{X = X^T \in {\mathbf R}^{D \times D}} f(X) 
	= \sum_{i,j=1}^D g([{\cal P}({\rm softplus}(X))]_{ij}), 
\end{eqnarray}

{\bf Step 3} Solve \eqref{unconstrained problem} 
through any stochastic (sub-gradient) descent method where 
the full sub-gradient of $f$ is approximated by 
\begin{equation}
	\label{gradient approximation}
	\nabla f(X) \approx \nabla g([{\cal P}
	({\rm softplus}(X))]_{i_*, j_*}), 
\end{equation}
with new pairs of indexes, $i_*,j_*\in \{1, \dots, D\}$, 
chosen uniformly at random at each iteration.

\paragraph{Remark}
Imposing \eqref{metric properties} or solving 
an all-pairs shortest path problem are computationally  
equivalent tasks (both scale as $O(D^3)$).
Through the stochastic approximation, however, 
the intrinsic-metric approach 
may reduce the complexity of the whole optimization
by up to an order of magnitude.
This because \eqref{gradient approximation} 
only requires the computation of a single-pair shortest-path, 
whose complexity is much lower than $O(D^3)$. 
On the other side, inequality-constrained optimization algorithms, 
which often are also iterative, need to impose
the full set of inequality constraints after each parameters update 
(as the solution is pushed back to the feasible 
set at each iteration).

\subsection{Algorithms}
For simplicity, we focus on a linear objective function 
\footnote{
A regularized version of $\ell(\alpha)$ is used in Sections 
\ref{section experiment 2} and \ref{section experiment 3} 
to train a ($k=1$) Nearest Neighbor algorithm based on
${\cal P}({\rm subplus}(M_{\alpha}))$.
	}
\begin{eqnarray}
	\label{objective linear}
	\ell(\alpha) = \ell(\alpha, \{M_r\}_{r=1}^R, Y) 
	= \sum_{i,j=1}^D (-)^{{\bf 1}_{y_i \neq y_j}} 
	[{\cal P}({\rm softplus}(M_\alpha))]_{ij}, 
\end{eqnarray}
where $M_\alpha = \sum_{r=1}^R \alpha_r M_{r}$, 
$M_r \in {\mathbf R}_+^{D \times D}$ and $M_r = M_r^T$ ($r=1, \dots, R$) 
are given input metrics
\footnote{Note that we do not require $M_r \in {\cal M}$. 
}
and $Y \in \{1, \dots, N_{\rm labels} \}^D$ is a given 
set of lables associated with the graph nodes.
On this case, we are able to prove the following properties 
of the projection method 
described in Section \ref{section metric constrained optimization} above.

\begin{lemma}
	$\ell(\alpha):{\mathbf R}^R \to {\mathbf R}$ is a continuous function.
\end{lemma}
{\bf Proof}: As 
$[{\cal P}({\rm softplus}(M_\alpha)))]_{ij} = 
\min_{\gamma \in \Gamma_{ij}} \gamma^T {\rm softplus}(m_\alpha)
$ ($\Gamma_{ij} = \Gamma_{ij}({\rm softplus}(M_\alpha))$ 
and $i,j=1, \dots, D$), the lemma follows directly 
from the continuity of the point wise minimum of a finite set of 
continuous functions. 
$\square$

\begin{theorem}
	\label{theorem subgradient}
	Let $\alpha_0 \in {\mathbf R}^R$ be a point where 
	\eqref{objective linear} is differentiable
	\footnote{\eqref{objective linear} is differentiable 
	almost everywhere.
	Non differentiable points correspond to $\alpha$ 
	such that there exists at least a pair of nodes for
	which the shortest path connecting them is not unique.
	}
	and let  
	${\cal U}_{\alpha_0}$ be a neighbourhood of $\alpha_0$ 
	such that for any $\alpha \in {\cal U}_{\alpha_0}$ and 
	any $i,j=1, \dots, D$ the shortest path 
	between nodes $i$ and $j$ computed on 
	${\cal G}({\rm softplus}(M_\alpha))$ 
	and ${\cal G}({\rm softplus}(M_{\alpha_0}))$ coincide.
	Sub-gradient, $V_{ij}^-$, and super-gradient 
	$V_{ij}^+$ of $[{\cal P}({\rm softplus}(M_\alpha)))]_{ij}$
	on ${\cal U}_{\alpha_0}$
	are given by 
	\begin{equation}
		\label{subgradient}
		\left[V^{\pm}_{ij}\right]_r = \gamma_{*}^T 
		(\sigma(m^{\pm})  \circ m_r), \quad 
		[m^-, m^+] = [m_{\alpha_0}, m_\alpha],
	\end{equation}
	where $r=1, \dots, R$ and   
	$\gamma_{*} = {\rm arg}\min_{\gamma \in \Gamma_{ij}} 
	\gamma^T {\rm softplus}(m_\alpha)$, with 
	$\Gamma_{ij} = \Gamma_{ij}({\cal G}({\rm softplus}(M_\alpha)))$.
\end{theorem}
{\bf Proof}: 
Let 
$g(x) = [{\cal P}({\rm softplus}(M_x))]_{ij} 
= \gamma_{x}^T {\rm softplus}(m_x)$ ($x = \alpha, \alpha_0$).
To prove that $V^-_{ij}$ is a sub-gradient of $g$, 
we need to show that $g(\alpha) - g(\alpha_0) 
\geq V_{ij}^T (\alpha - \alpha_0)$.
The linearity of $V_{ij}$ let us rewrite the right-hand side 
of the inequality as 
$\gamma^T_{\alpha_0} \sigma(m_{\alpha_0}) 
\circ (m_{\alpha} - m_{\alpha_0})$. 
On ${\cal U}_{\alpha_0}$, the left-hand 
side can be rewritten as   
$\gamma_{\alpha_0}^T({\rm softplus}(m_\alpha) - {\rm softplus}(m_{\alpha_0}))$
From the mean value theorem we obtain 
$g(\alpha) - g(\alpha_0) = 
\gamma_{\alpha_0}^T (\sigma(m_{\bar \alpha}) 
\circ (m_{\alpha} - m_{\alpha_0}))$, 
where $m_{\bar \alpha} = (1 - t) m_{\alpha} + t m_{\alpha_0}$.
Finally, $\gamma_x \geq 0$ implies 
$\gamma^T_{\alpha_0} \sigma(m_{\alpha}) \circ (m_{\alpha} - m_{\alpha_0})
\geq 
\gamma^T_{\alpha_0} \sigma(m_{\alpha_0}) \circ (m_{\alpha} - m_{\alpha_0})$
since, for each $I=1, \dots, D^2$, we obtain 
$(\sigma(\bar z) - \sigma(z_0)) (z - z_0)\geq 0$, 
where $z = [m_{\alpha}]_I$ (idem $\bar z$ and $z_0$) and 
$\bar z \in [z, z_0]$.
This is true because $\sigma(x)$ 
is a monotone increasing function of its argument.
The proof for $V^+_{ij}$ is analogous.
$\square$

\begin{lemma}
	\label{subgradient ell}
	Let $\alpha_0 \in {\mathbf R}^R$ and 
	${\cal U}_{\alpha_0}$ be the neighborhood of $\alpha_0$ defined in 
	Theorem \ref{theorem subgradient}.
	For $\alpha \in {\cal U}_{\alpha_0}$,  
	the sub-gradient of $\ell(\alpha)$ is 
	defined by 
	\begin{eqnarray}
		\label{full subgradient}
		\sum_{i,j=1}^D ([C_+]_{ij} [V^{-}_{ij}]_r 
		- [C_-]_{ij} [V_{ij}^{+}]_r 
	\end{eqnarray}
	where $r=1, \dots, R$, $i,j=1, \dots, D$, 
	$V_{ij}^{\pm}$ are the sub- and super-gradients defined in 
	Theorem \ref{theorem subgradient},  
	$[C_+]_{ij} = {\bf 1}_{y_i = y_j}$ and
	$[C_-]_{ij}= {\bf 1}_{y_i \neq y_j}$.
\end{lemma}
{\bf Proof}: Let $x = {\cal P}({\rm softplus}(m_\alpha))$, then 
$\ell(\alpha) = \ell(x) = c_+^T x - c_-^T x$ and 
$\ell(x) - \ell(x_0) = 
(c_+ - c_-)^T (x - x_0)
\geq \sum_{r = 1}^R (c_+^T [v^{-}]_r - c_-^T [v^{+}]_r) (\alpha - \alpha_0)$.
$\square$

\paragraph{Remark}
Sub- and super-gradients in \eqref{subgradient} 
could be also obtained from a 
smooth approximation of the non-differentiable `min' function 
in \eqref{projector}.
For example, let 
$\min_\gamma \gamma^T {\rm softplus}(m_\alpha) 
= \lim_{T \to \infty} \tilde \ell(\alpha, T)$, 
with $
\tilde \ell(\alpha, T)
=  T^{-1} \log(\sum_\gamma 
e^{- T \gamma^T {\rm softplus}(m_\alpha)})$.
When $\ell(\alpha)$ is differentiable, its gradient 
can be obtained by exchanging the limit and the 
derivative in 
$\partial_\alpha \lim_{T\to \infty} \tilde \ell(\alpha, T)$. 
To obtain a formal proof, if it is necessary to 
show the absolute convergence of $\tilde \ell(\alpha) \to \ell(\alpha)$ 
on some intervals of ${\mathbf R}^R$.
\footnote{\cite{pillutla2018smoother} contains an interesting analysis 
of how smooth versions of the max function can help 
structured machine learning algorithms.}
This is the general idea we use to prove Theorem \ref{theorem gradient}.
We first need the following lemma: 
\begin{lemma}
	\label{lemma sum exp convergence}
	Let $g(t, x, {\cal F} ) = 
	- \frac{1}{t} \log{\left( \sum_{k=1}^K e^{- t f_{k}(x)}\right)}$, 
	where ${\cal F} = \{f_k(x):{\mathbf R}^R \to {\mathbf R} \}_{k=1}^K$ 
	is a set of continuous and differentiable functions and $t > 0$.
	Then 	
	\begin{eqnarray}
		\lim_{t \to \infty} g(t, x, {\cal F})  &=&
		\min_{k} f_k(x) \\
		\lim_{t \to \infty} [\nabla g]_{r} &=&
		[\nabla f_{\bar k}]_r, \quad \bar k = {\rm arg} 
		\min_{k} f_k(x) 
	\end{eqnarray}
	and the convergence is uniform 
	on all intervals $I\subset{\mathbf R}^R$ such that 
	$f_{k}(x) \neq f_{k'}(x)$, for all 
	$k \neq k'$ ($k,k'=1, \dots, K$) and all $x \in I $, and 
	$|f_{k}(x)| < \infty$ and 
	$\|\nabla f_{k}\| < \infty$ for all $k = 1, \dots, K$ and all 
	$x \in I $.
\end{lemma}
{\bf Proof}:
A sequence of functions $\{g_{n}: I \to {\mathbf R}\}_{n=1}^\infty$
converges uniformely to $g : I \to {\mathbf R}$
if $\lim_{n \to \infty} g_n(x) = g(x)$ for every $x \in I$ 
(point wise convergence) 
and, for all $n=1, \dots, \infty$, there exists $M_n < \infty$ such that
$|g_{n}(x)| < M_n$, for all $x \in I$, and the sequence 
$\{ M_n\}_{n=1}^\infty$ converges (uniform convergence).
Let $\{ t_n \in {\mathbf R}\}_{n=1}^{\infty}$ be a sequence of real numbers 
such that $t_{n} < t_{n+1}$ and $\lim_{n \to \infty} t_n = \infty$ and 
$g_n(x) = g(t_n, x, {\cal F})$. 
Then, for any $x \in I$,  
\begin{eqnarray}
	\lim_{n \to \infty} g_n(x) & = & 
	 - \lim_{n \to \infty} \frac{1}{t_n} 
	\log{\left( \sum_{k=1}^K e^{- t_n f_k(x)}\right)}\\
	& = & f_{\bar k}(x) -  \lim_{n \to \infty} \frac{1}{t_n} 
	\log{\left( 1 + \sum_{k=1}^K 
	e^{- t_n (f_k(x) - f_{\bar k}(x))}\right)}\\
	& = & f_{\bar k}(x) 
\end{eqnarray}
where $\bar k = {\rm arg} \min_{k} f_{k}(x)$. 

The convergence is uniform because, for example,  
$M_n = \max_{x \in I} |\min_{k} f_{k}(x)| + \frac{1 + K}{t_n}$ is such that
$\lim_{n \to \infty} M_n < \infty$ and, for all $n=1, \dots, \infty$ and 
all $x \in I$, 
\begin{eqnarray}
	|g_n(x)| & = & \left| 
	f_{\bar k}(x) -  \frac{1}{t_n} 
	\log{\left( 1 + \sum_{k=1}^K 
	e^{- t_n (f_k(x) - f_{\bar k}(x))}\right)} 
	\right| \\
	& \leq &  
	|f_{\bar k}(x)| +  \frac{1}{t_n} 
	\log{\left( 1 + \sum_{k=1}^K 
	e^{- t_n (f_k(x) - f_{\bar k}(x))}\right)} \\
	& \leq &  
	\max_{x \in I} |\min_{k} f_{k}(x)|  +  \frac{1 + K}{t_n} = M_n
\end{eqnarray}
When 
$\min_{x} \min_{k \neq k'} |f_{k}(x) - f_{k'}(x)| > 0$, 
and $\| \nabla f_{k}\| < \infty$ for all $k = 1, \dots, K$ 
and all $x \in I$,
we also have 
\begin{eqnarray}
	\lim_{n \to \infty} [\nabla g_n]_r & = & 
	 \lim_{n \to \infty} 
	 \frac{\sum_{k=1}^K [\nabla f_{k}]_r e^{- t_n f_k(x)}}
	 {\sum_{k=1}^K e^{- t_n f_k(x)}}
	 \\
	 & = &\lim_{n \to \infty}\frac{
		[\nabla f_{\bar k}]_r + 
		\sum_{k \neq \bar k}^K [\nabla f_{k}]_r
	 e^{- t_n (f_k(x) - f_{\bar k}(x))}} 
	 {1 + \sum_{k \neq \bar k}^K e^{- t_n (f_k(x) - f_{\bar k}(x))}} \\
	  & = &  [\nabla f_{\bar k}]
\end{eqnarray}
for all $r=1, \dots, R$, since all extra terms in the numerator and denominator vanish 
for $t_n \to \infty$.
This happens because $f_{k}(x) - f_{\bar k}(x) > 0$ for all $k\neq \bar k$ and all $x \in I$. 
To prove that the convergence is uniform we let
$M_n = \max_{x \in I} |[\nabla f_{\bar k}]_r| 
+ K C e^{- t_n \Delta}$, 
where $C = \max_{k} |[\nabla f_{k}]_r | < \infty$ and 
$\Delta = \min_{x \in I} \min_{k \neq k'} |f_{k}(x) - f_{k'}(s)| > 0$. 
Then it is easy to show that
\begin{eqnarray}
	|\nabla g_n| < M_n, \quad {\rm for \ all \ } n=1, \dots, \infty, \quad {\rm and} \quad \lim_{n \to \infty} M_n < \infty, 
\end{eqnarray}
which completes the proof. $\square$

\begin{theorem}[{\bf Theorem 7.17 of \cite{rudin1964principles}}]
	\label{theorem rudin}
	Let $\{g_n\}_{n=1}^\infty$ be a sequence of functions that converges 
	uniformely to $g$ on $I \subset {\bf R}^R$.
	If $\{[\nabla g_n]_{r}\}_{n=1}^\infty$ converges uniformely 
	to $[\nabla g]_r$, then
	\begin{equation}
		[\nabla g]_{r} = \lim_{n \to \infty} [\nabla g_n]_r
	\end{equation}
\end{theorem}
{\bf Proof}: See \cite{rudin1964principles} $\square$

\begin{theorem}
	\label{theorem gradient}
	Assume $0 < [M_{r}]_{ij} < \infty$ for all $i,j=1, \dots, D$ and 
	all $r=1, \dots, R$.
	Let $I \subset {\mathbf R}^{R}$ be such that the 
	shortest path between nodes $i$ and $j$ 
	on graph ${\cal G}({\rm softplus}(M_\alpha))$ is unique 
	for all $\alpha \in I$ and all $i, j=1, \dots, D$.
	Then 
	\begin{equation}
		\left[ \nabla [{\cal P}({\rm softplus}(M_\alpha))]_{ij} 
		\right]_r = \gamma_{*}^T (\sigma(m_{\alpha}) \circ m_r) 
	\end{equation}
	where $\alpha \in I$, $i,j = 1, \dots, D$, $r=1, \dots, R$, 
	$\gamma_{*} = {\rm arg}\min_{\gamma \in \Gamma_{ij}}
	\gamma^T {\rm softplus}(m_\alpha)$
	and 
	$\Gamma_{ij} = \Gamma_{ij}({\cal G}({\rm softplus}(M_\alpha)))$.
\end{theorem}
{\bf Proof}: The statement follows directly 
from Theorem \ref{theorem rudin} and Lemma \ref{lemma sum exp convergence} 
by letting $f_k(\alpha) = \gamma_k^T {\rm softplus}(m_\alpha)$, 
$\alpha \in I$,
$\gamma_{k} \in  \Gamma_{ij}(\cal G)$ and 
${\cal G}({\rm softplus}(M_\alpha))$
for all $i,j=1, \dots, D$. 
It is easy to check that such $\{f_k(\alpha)\}_{k=1}^{K}$ fulfil 
the assumption of Lemma \ref{lemma sum exp convergence} when $I \subset {\mathbf R}^R$ is an 
interval where the shortest path between nodes $i$ and $j$ on 
${\cal G}({\rm softplus}(M_\alpha))$, 
$\gamma_{*} = {\rm arg}\min_{\gamma \in \Gamma_{ij}}
\gamma^T {\rm softplus}(m_\alpha))$, is unique
for all $i,j=1, \dots, D$ and when  
$0 < [M_{r}]_{ij} < \infty$ for all $i,j=1, \dots, D$ and 
all $r=1, \dots, R$.
$\square$

\begin{figure}[h!]
\vskip 0.2in
\begin{center}
	\includegraphics[width=.4\columnwidth]{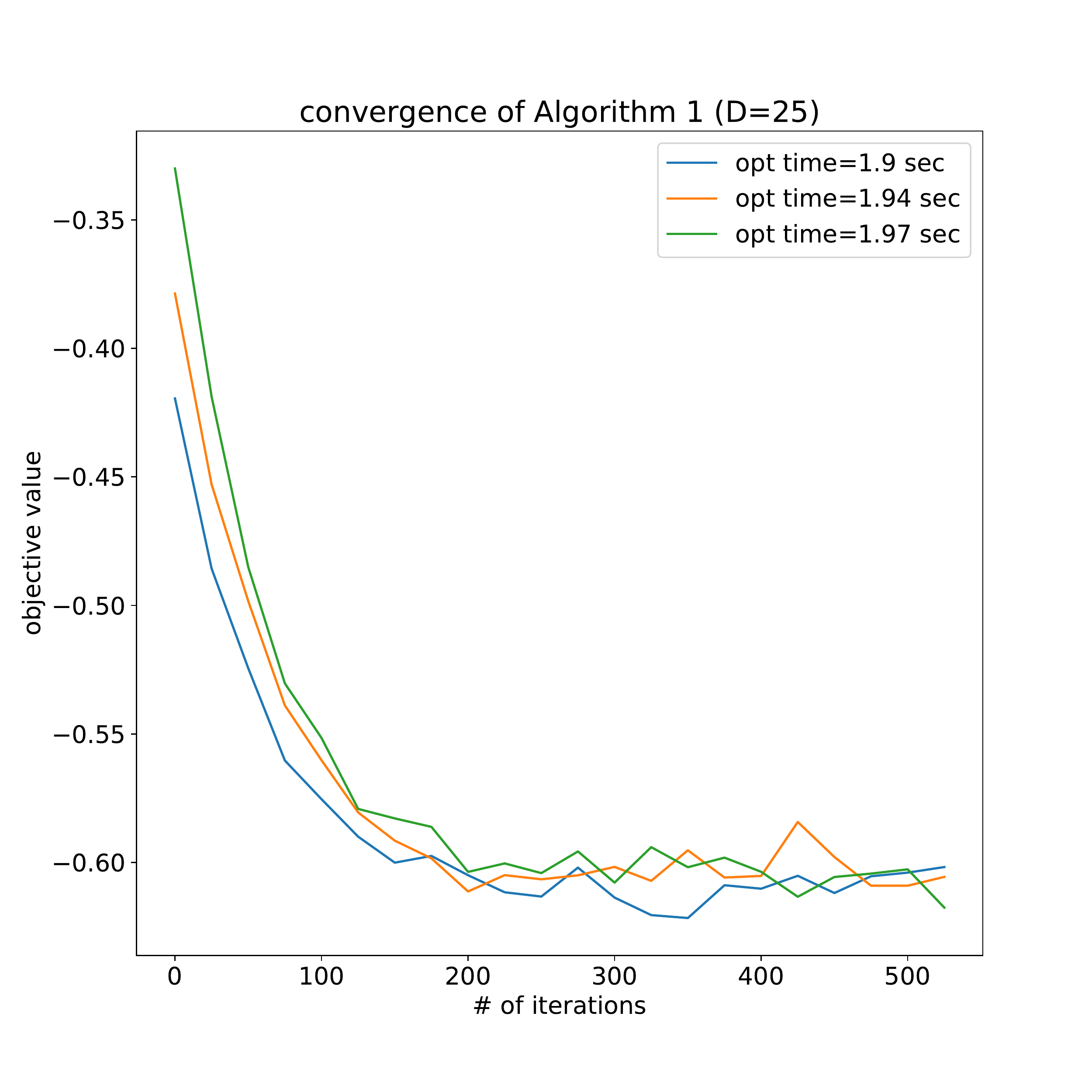}
	\includegraphics[width=.4\columnwidth]{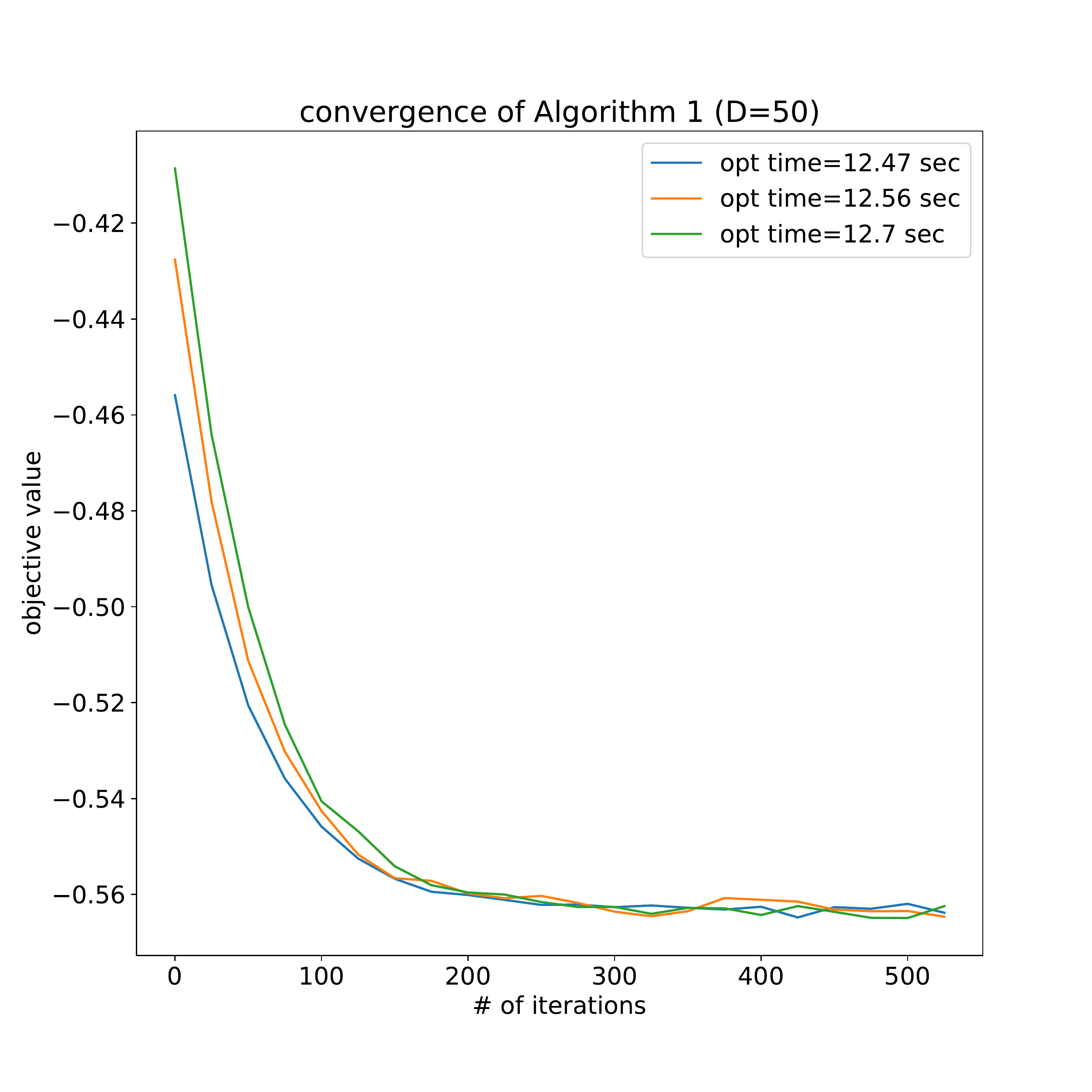}
	\\
	\includegraphics[width=.4\columnwidth]{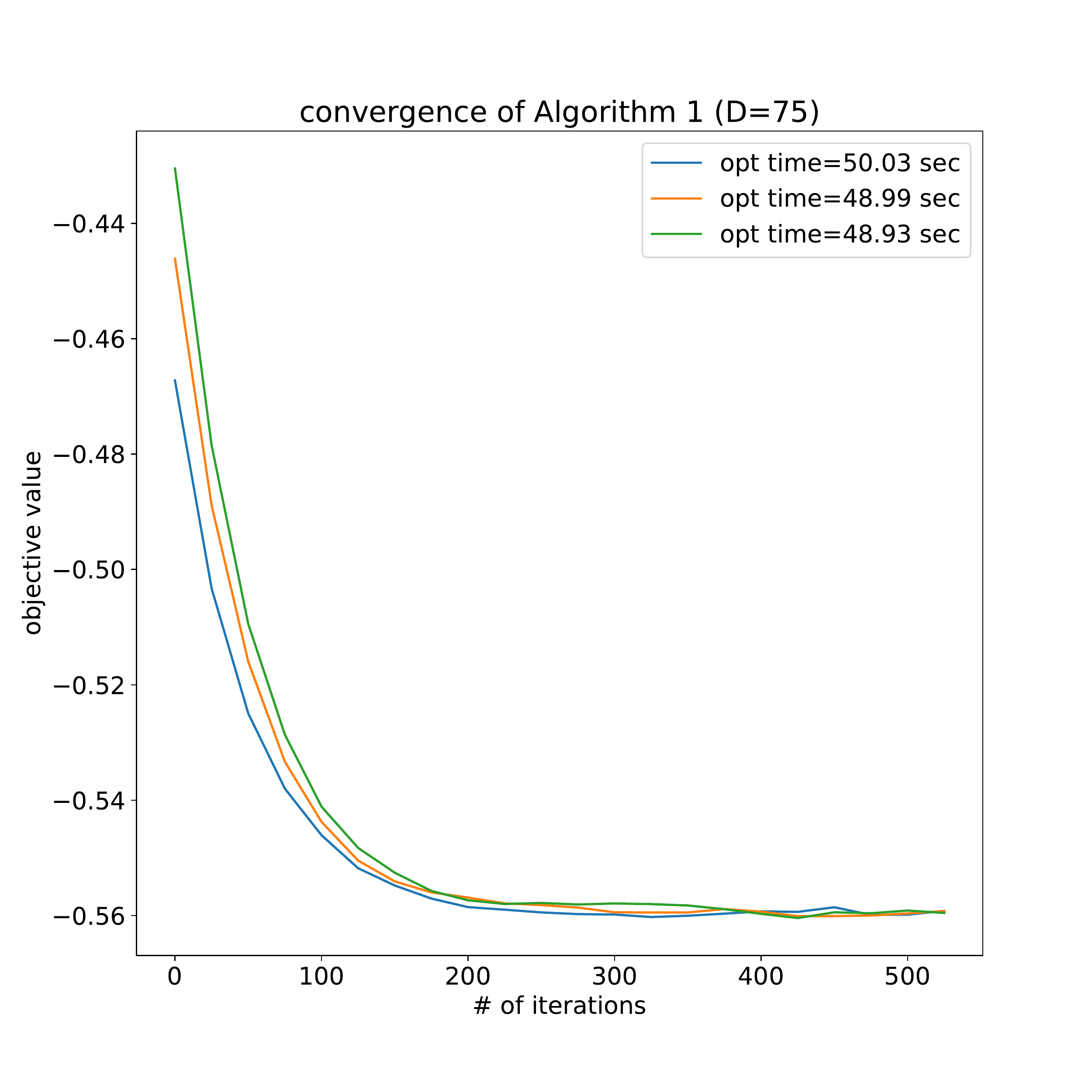}
	\includegraphics[width=.4\columnwidth]{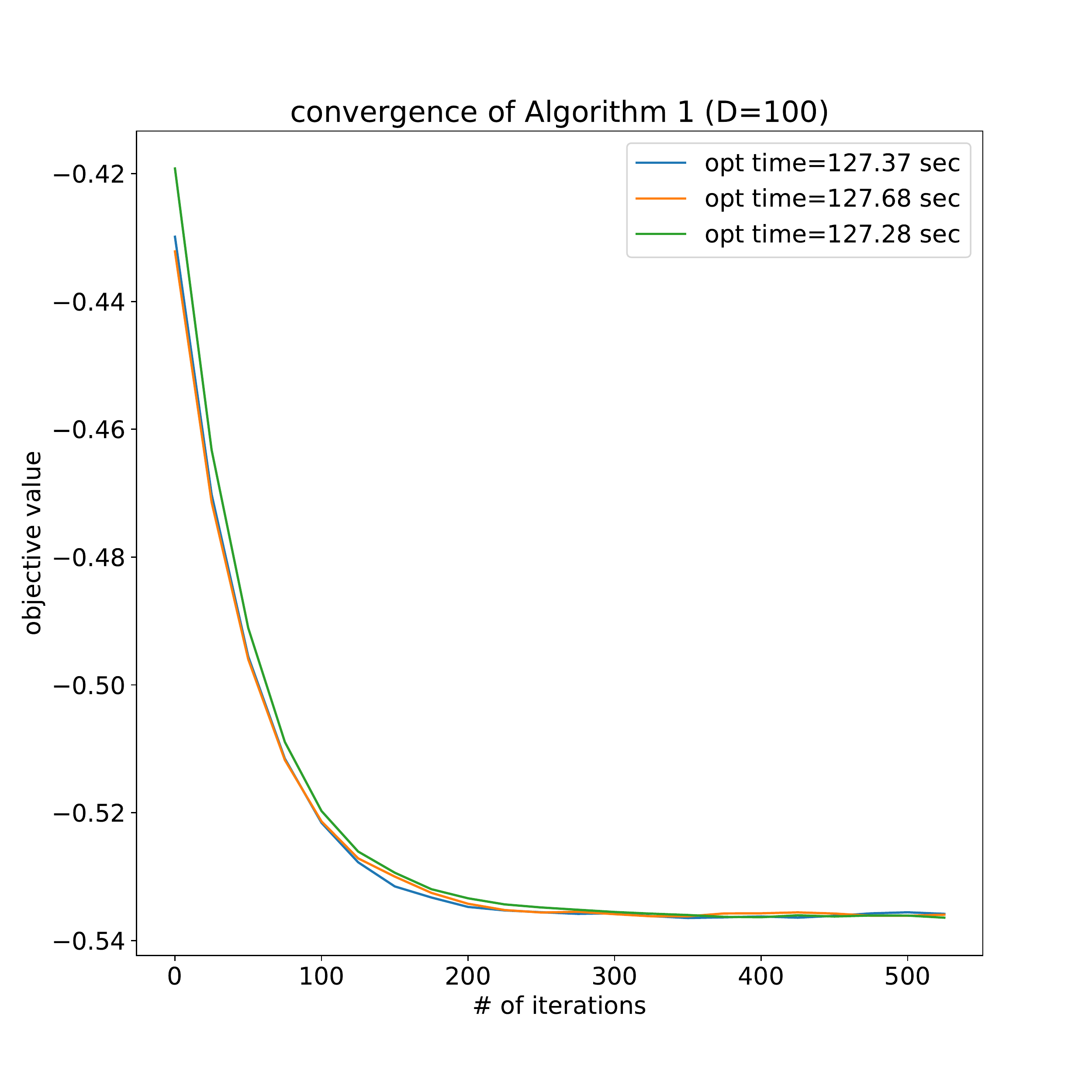}
	\caption{Convergence of Algorithm \ref{algorithm 1} from 
	slightly different initializations 
	$\alpha  = \frac{x}{\| x \|^2}$, with $x \sim {\cal N}(0, 1)^R$ and $R = 8$, and different 
	size, $D$, of the input metrics $M_{r}$ ($r = 1, \dots, R$, see 
	Section \ref{section experiment 2} for more details). 
	On the $y$-axis, we plot the value of 
	$\ell(\alpha^{(k)}) + \rho \| \alpha^{(k)}\|^2$ ($\rho = 0.01$)
	evaluated on the training set.
	For all runs, we use the same data set, obtained from the MNIST 
	database as explained in Section \ref{section experiment 2}, and 
	same learning parameters, $\eta = 1$ and $k_{\max} = 500$.
	}
\label{figure algorithm convergence}
\end{center}
\vskip -0.2in
\end{figure}

\paragraph{Stochastic sub-gradient descent}
\eqref{full subgradient} and standard stochastic 
sub-gradient descent algorithms.\footnote{
\cite{shor2012minimization} is a classical reference for sub-gradient 
methods and their convergence properties (see also \cite{boyd2003subgradient}).
More details and properties of their stochastic version can be found in 
\cite{shor2013nondifferentiable}
} can then be used to minimize \eqref{objective linear} 
or its norm-regularized 
version used in the experiments of  
Sections \ref{section experiment 2} and \ref{section experiment 3}.
Algorithm \ref{algorithm 1} describes 
the proposed optimization methods for this special case, 
i.e. to solve 
\begin{eqnarray}
	\label{optimization algorithm 1}
	{\rm minimize} & \ell(\alpha) = \ell(M_{\alpha}, \{M_r \}_{r=1}^R, Y) \\
	{\rm s.t.} & M_{\alpha} \in {\cal M}
\end{eqnarray}
where
$\ell(\alpha) = 
D^{-2} \sum_{i,j=1}^D 	(-)^{y_i \neq y_j} + \rho \| \alpha\|^2$ with
$M_r \in {\mathbf R}_+^{D \times D}$ and $M_r = M_r^T$ ($r=1, \dots, R$), 
$Y \in \{1, \dots, N_{\rm labels} \}^D$ and 
$\rho > 0$. 
Differently from usual sub-gradient descent algorithms, 
we do not check if the sub-gradient is a descent direction at 
each iteration.
Such a check is based on the evaluation of the full-sample objective
and, in our settings, requires solving 
the solution of an expansive all-pairs shortest 
path problem. 
To preserve the computational advantages of our 
approach, we use instead unchecked updates 
$\alpha^{(k + 1)} = \alpha^{(k)} - \eta g^{(k)}$.
Figure \ref{figure algorithm convergence} shows that  
Algorithm \ref{algorithm 1} retains good convergence 
properties even with unchecked updates.
Another option (not implemented here) would be to evaluate the objective 
on an arbitrary small validation data set extracted from 
the training samples. 	
This would also allow a formal convergence proof 
and the definition of early stopping criteria.
In Algorithm \ref{algorithm 1}, we also implicitly 
assume that $\ell(\alpha^{(k)})$ is differentiable for all $k$ and 
use a simplified expression for its sub-gradients.
If $\ell(\alpha)$ is differentiable in $\alpha$ it is possible 
to choose $\alpha = \alpha_0$ in \eqref{subgradient} so that 
sub- and super-gradient coincide with the true gradient, 
provided by Theorem \ref{theorem gradient}.

\begin{algorithm}
	\caption{Soution of \eqref{optimization algorithm 1}.
	}
\label{algorithm 1}
	\begin{algorithmic}[1]
	\STATE {\bf Input:}
	input metrics $\{M_r\}_{r = 1}^R$, 
	training labels $\{ (y^{(d)} \}_{d =1}^D$, 
	initialization $\alpha_0 \in {\mathbf R}^{R}$, 
	step size $\eta > 0$, 
	maximum number of iterations $k_{\rm max} \in {\mathbf N}_+$, 
	regularization parameter $\rho > 0$
   	\STATE Initialize $\alpha^{(0)} = \alpha_0$
	\WHILE{$k < k_{\rm max}$}
		\STATE sample $(i, j)$ uniformly at random 
		in $\{1, \dots, D\} \times \{1, \dots, D\} $
		\STATE let 
		${\cal G} = {\cal G}(M_{\alpha^{k}})$
		\STATE let 
		$\gamma_*  = \min_{\gamma \in \Gamma_{ij}({\cal G})} 
		\gamma^T{\rm softplus}(m_\alpha^{(k)})$
		\FOR{$r = 1, \dots, R$}
			\STATE let 
			$[{\rm grad}^{(k)}]_r = D^{-2}(-)^{{\bf 1}_{y_i \neq y_j}}
			\gamma_*^T (\sigma([m_{\alpha^{(k)}}]) \circ [m_r])$
		\ENDFOR
		\STATE let 
		$\alpha^{(k + 1)} = \alpha^{(k)} 
		- \eta ({\rm grad}^{(k)} + 2 D^{-2} \rho \alpha^{(k)})$
	\ENDWHILE	
	\STATE {\bf Output:} $\alpha_* = \alpha^{(k_{\rm max})}$
\end{algorithmic}
\end{algorithm}

\section{Experiments}
\label{section experiments}
\subsection{Data}
We use two real-world data sets: 
i) the MNIST data set, consisting of 
$N = 70000$ images of grey-scale hand-written digits \cite{lecun1998gradient}, and 
ii) the Citeseer dataset, containing $N = 3327$ scientific papers 
and the corresponding citation network \cite{sen2008collective}.
\paragraph{MNIST data}
We pre-process all grey-scale images in the MNIST data set 
and extract feature vectors of $d_x = 784$ strictly positive entries.
The images are labelled with integers $n=0, \dots, 9$.

\begin{figure}[h!]
\vskip 0.2in
\begin{center}
	\includegraphics[width=.4\columnwidth]{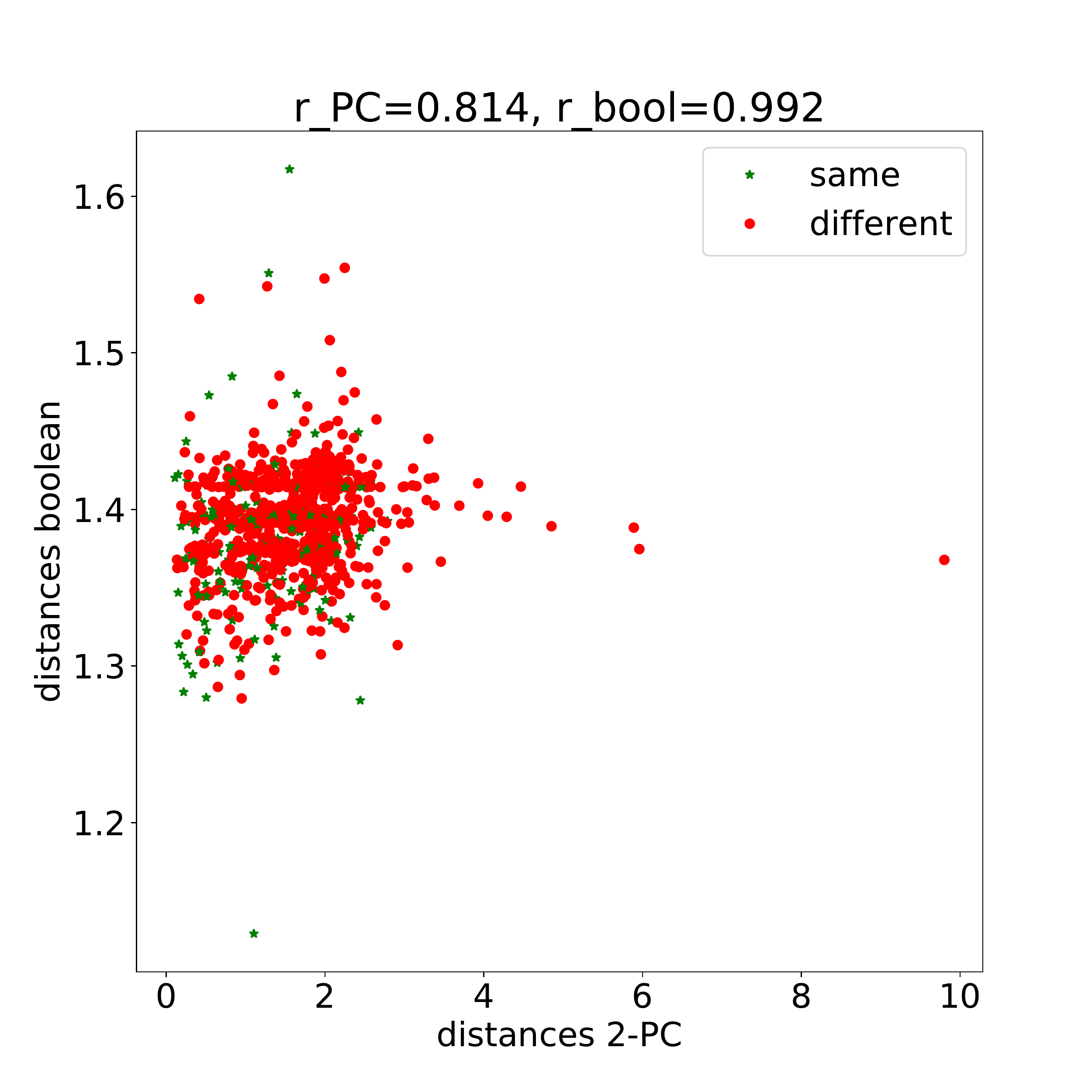}
	\includegraphics[width=.4\columnwidth]{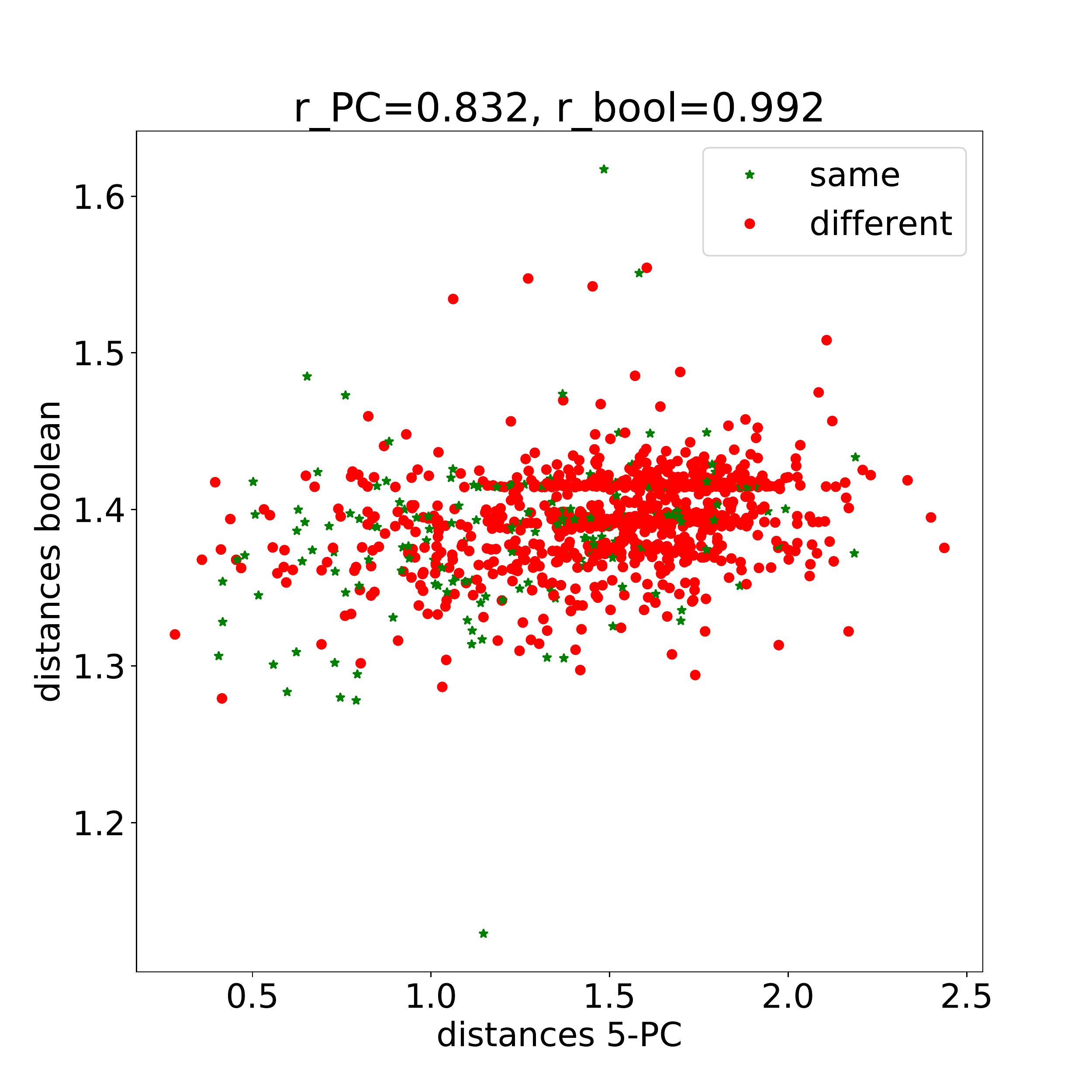}\\
	\includegraphics[width=.4\columnwidth]{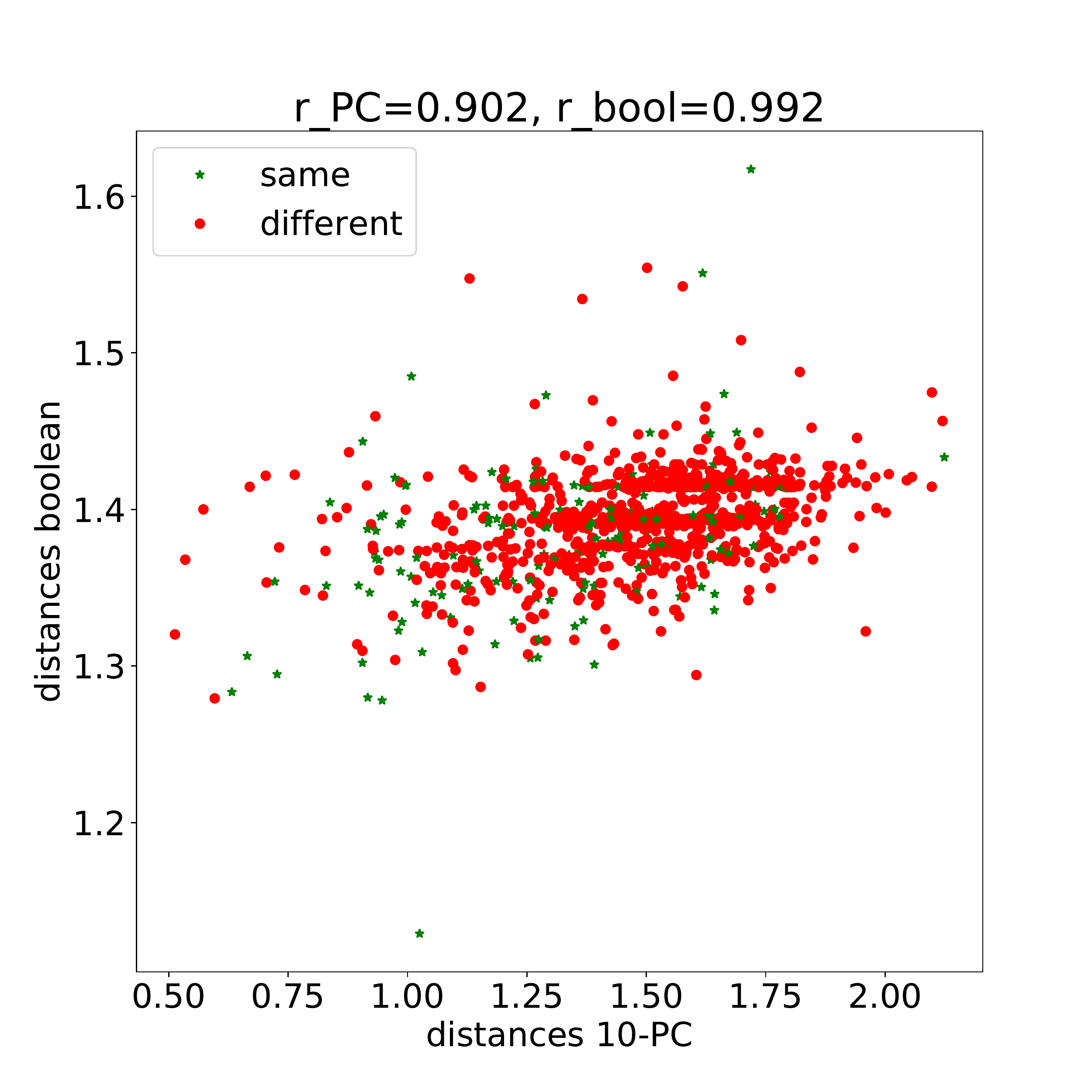}
	\includegraphics[width=.4\columnwidth]{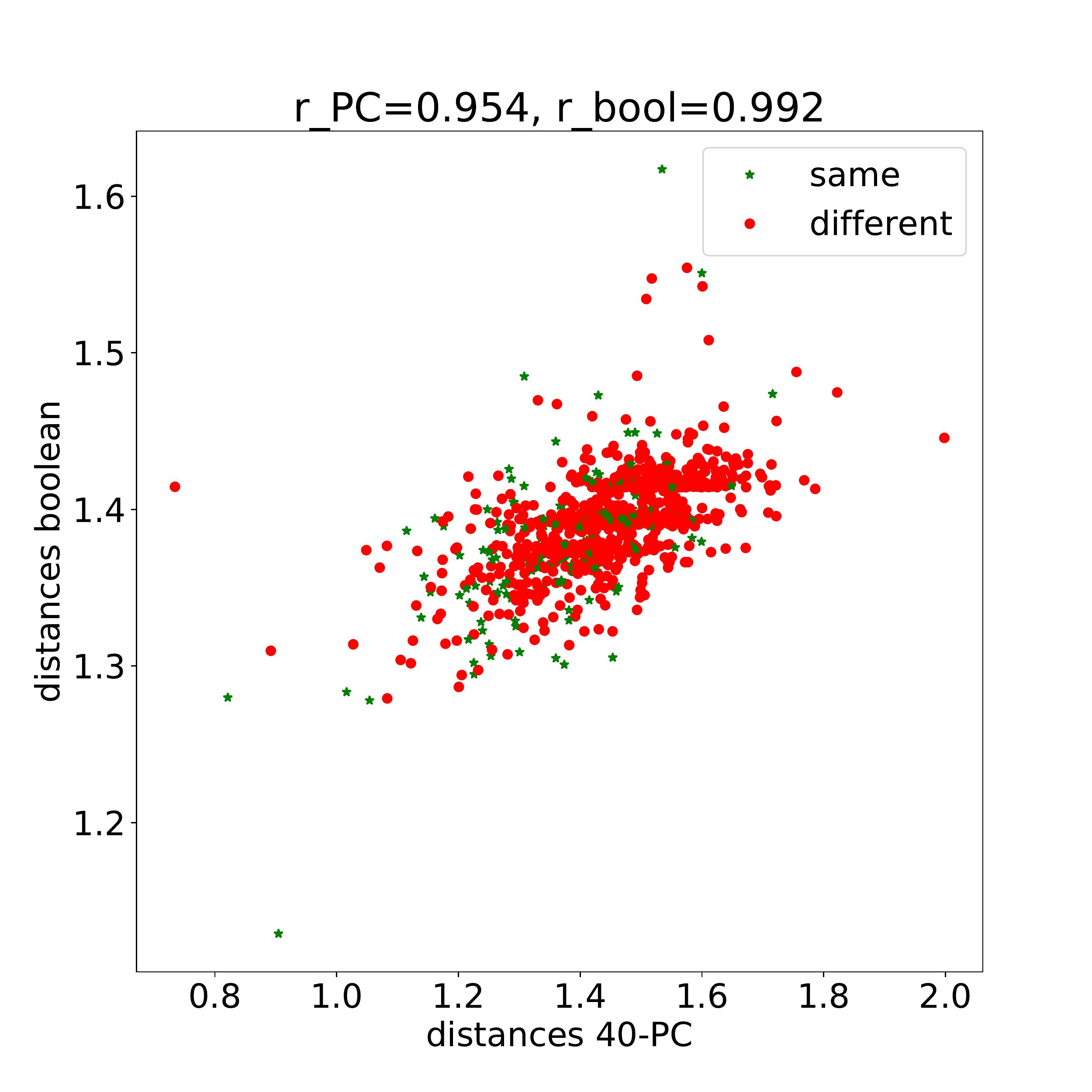}
	\caption{Cluster separation of the original (Boolean) word-embedding
	vectors and their projection on $n=2,5, 10, 40$ principal components.
	For each projected data set, $X_{n \ {\rm PC}}$ ($n=2, 5, 10, 40$), 
	and each paper pair (we randomly select 1000 paper pairs), 
	we plot the distance between the corresponding 
	word-embedding ($y$-axis) and PC vectors ($x$-axis).
	Different markers are used for pairs picked from 
	the same cluster or different clusters.
	The total separation score, $r_{\rm data}$ (
	${\rm data} \in \{ n \ {\rm PC}, bool \} $) 
	is the ratio between the average distance
	between papers in the same cluster, {\bf same}, and between 
	papers in different clusters, {\bf different}.
	} 
\label{figure preliminary test}
\end{center}
\vskip -0.2in
\end{figure}

\begin{figure}[h!]
\vskip 0.2in
\begin{center}
	\includegraphics[width=.8\columnwidth]{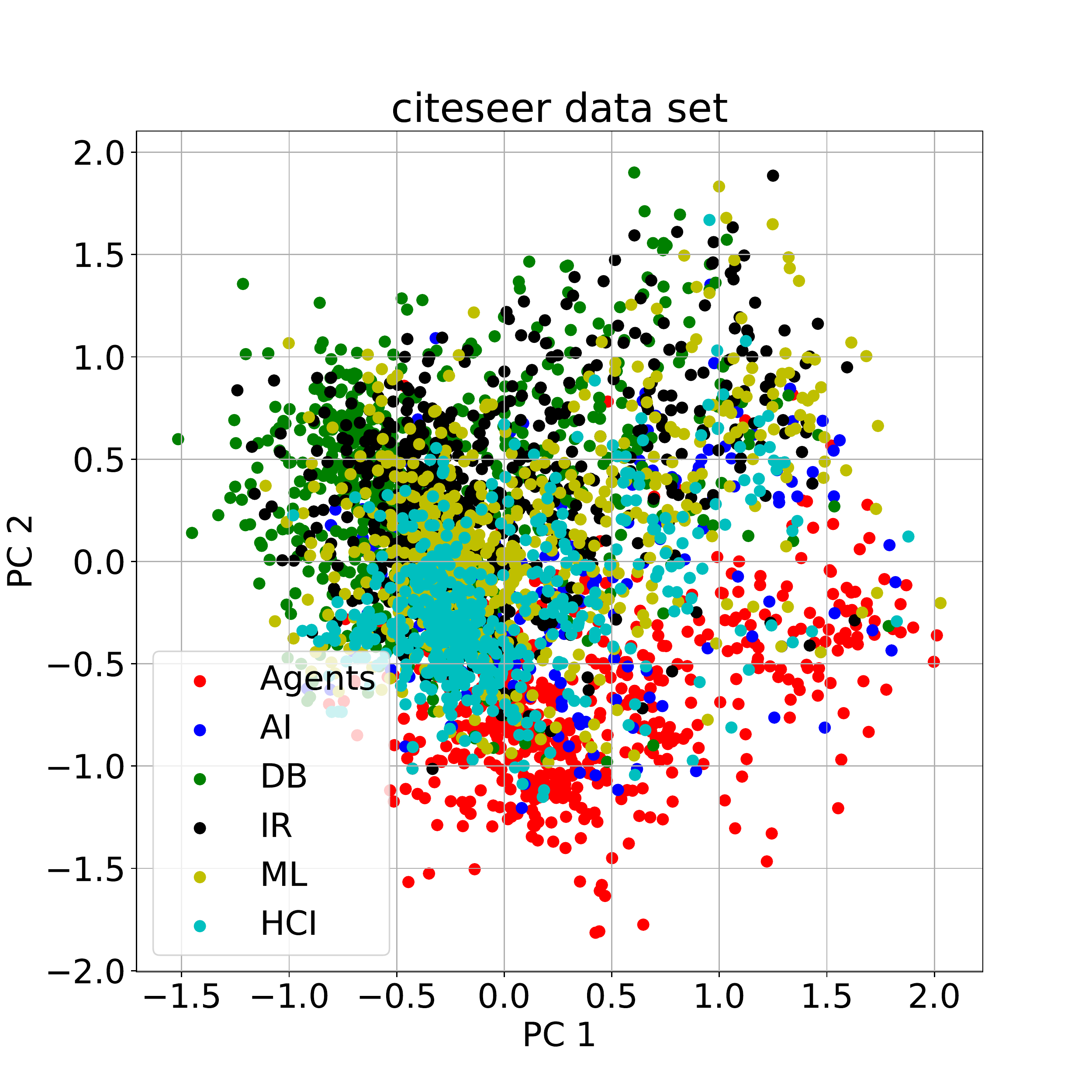}
	\caption{Two-dimensional visualization of the Citeseer data set.
	Dimensional reduction is obtained by projecting on the first and 
	second principal components.
	Different colors correspond to different topics.
	} 
\label{figure citeseer data}
\end{center}
\vskip -0.2in
\end{figure}

\paragraph{Citeseer data}
The Citeseer data set is a widely used as benchmark 
dataset for semi-supervised learning algorithms 
\cite{sen2008collective, kipf2016semi}.
The scientific papers are represented by 
bag-of-words feature vectors of dimensionality $d_x = 3703$ and 
assigned according to their topic to 6 non-overlapping 
classes.
The $d_x$-dimensional Boolean vectors are transformed into 
$40$-dimensional real vectors by projecting them onto the space of 
their first $40$ principal components.
PCA projections can be shown to increase the clusters separation, 
but this is an optional step and our method would apply
without changes with Boolean inputs. 
Figure \ref{figure preliminary test} shows 
the predictive power of 
the Boolean data set, $X_{\rm bool}$, 
and a set of projected data sets, $X_{n \ {\rm PC}}$ ($n=2, 5, 10, 40$).
For each data set, $X_{\rm PC} = X_{n \ {\rm PC}}$, we have compared the ratio 
$r_{\rm data} = \frac{v_{\rm same}(X)}{v_{\rm different}(X)}$, 
where
${\rm data} \in \{ PC, bool \}$
$v_{\rm same} = \sum_{(y, x), (y', x') \in X} 
{\bf 1}_{y =y'} \|x - x' \|^2$ and   
$v_{\rm diff} = \sum_{(y, x), (y', x')\in X} 
{\bf 1}_{y \neq y'} \|x - x' \|^2$ and
$X = X_{\rm bool}$ and $X = X_{\rm PC} \}$. 
Figure \ref{figure citeseer data} shows a two-dimensional (PCA) 
reduction of the data set.
To make the calssification task harder, in Section \ref{section experiment 3} 
we use $X = X_{40 \ {\rm PC}}$.

\subsection{Experiment 1: computational complexity (MNIST data)}
\label{section experiment 1}
\begin{figure}[h!]
\vskip 0.2in
\begin{center}
	\centerline{\includegraphics[width=\columnwidth]{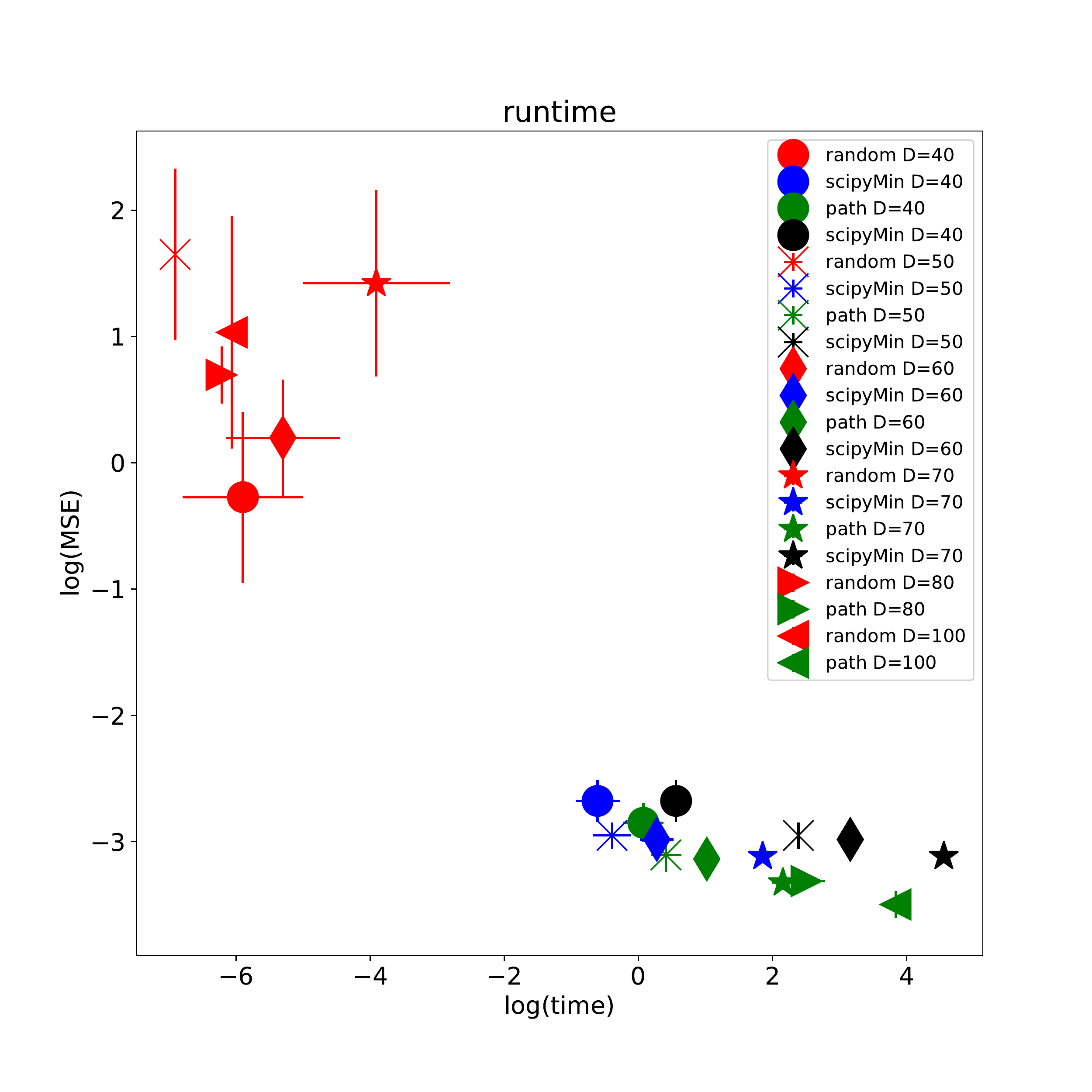}}
	\caption{Runtime versus Mean Square Error at convergence.
	For {\bf cvx}, values associated with 
	colored markers do not 
	include the computational cost of forming the
	constraints matrix, $A^{(D)}_{\rm constraints} \in \{ -1, 0, 1\}^{O(D^3) \times O(D^2)}$, 
	which is a required input of {\bf cvx}.
	The corresponding total runtime, i.e. the 
	optimization time plus the time needed to 
	compute $A_{\rm constraints}^{(D)}$,
	is indicated by black markers 
	and referred to by ${\bf cvx + extra}$ in the legend.
	Error bars represent standard deviations over 3 analogous 
	experiments.
	Markers {\bf cvx  D=80} and {\bf cvx D = 100} are not 
	present because memory errors occurred while computing 
	$A_{\rm constraints}^{(D)}$.
	}

\label{figure runtime}
\end{center}
\vskip -0.2in
\end{figure}

To show the computational complexity gain associated with 
the intrinsic-metric approach we compare the proposed 
algorithm with a direct method, where metric constraints 
are imposed explicitly. 
We consider the optimization problem 
\begin{eqnarray}
	\label{exp 1 optimization problem}
	\min_{\alpha \in {\mathbf R}^R} && \ell(\alpha) =  D^{-2} \|  
	  M_{\rm true}- M_{\alpha} \|^2 \\
	{\rm s.t. } && M_{\alpha} = \sum_{r=1}^R\alpha_r M_r \in {\cal M}, 
\end{eqnarray}
where $D$ is the number of images extracted from 
the MNIST data set and $M_{\rm true}$ and $M_r$ ($r = 1, \dots, R$) 
are metrics of the selected data set defined by 
\begin{eqnarray}
	\label{exp 1 metrics}
	\left[ M_{\rm true}\right]_{ij} & = & \frac{\| x^{(i)} - x^{(j)} \|}{\sqrt{\|x^{(i)}\| \| x^{(j)}\|}}, \\ 
	\left[ M_{r}\right]_{ij} &=& {\rm dist}_{\rm path}(i, j, G(A_r)), \\
	\left[ A_r \right]_{ij} &=& {\bf 1}_{[M_{\rm true}]_{ij} < \xi_r}, 
\end{eqnarray}
with 
$\xi_r = \bar a (\frac{1}{4} + (r -1) \frac{6}{4 R})$, 
$\bar a = \frac{1}{D^2} \sum_{i,j=1}^D [M_{\rm true}]_{ij}$ 
and $R = 8$.
We compare the performance of the proposed method, {\bf path}, 
with a state-of-the-art solver for 
constrained-quadratic programming, {\bf cvx}.
A regularization term, $\rho \| \alpha \|^2$ is added 
to objective minimized by {\bf cvx} to make improve 
convergence speed.
{\bf path} is obtained by adapting Algorithm \ref{algorithm 1} 
to the least-squares objective $\ell(\alpha)$ 
defined in \eqref{exp 1 optimization problem}.
The learning rate, $\eta$, is fixed for all $D$
and the optimization is 
stopped when $\ell^{(t+1)} < \ell^{(t)} (1 + .0001)$, 
where $\ell^{(t)} = \ell(\alpha^{(t)})$ is the objective function 
defined in 
\eqref{exp 1 optimization problem} evaluated on the training 
set and $\alpha^{(t)}$ is the 
value of the model parameter at the $t$th iteration.
To assess the quality of the algorithms' output, we consider a baseline, 
{\bf rand}, where 
$\alpha_{\rm rand} = \frac{x}{\| x\|}$, with $x \sim {\cal N}(0, 1)^R$.
To facilitate the comparison and make the optimization 
more stable, we normalize $M_r$  in such 
a way $\| M_{r} \|  = \| M_{\rm true} \|$ ($r=1, \dots, R$).
For different sizes of the input metrics, $D = 40, 50, 60, 70, 80, 100$, 
Figure \ref{figure runtime} shows 
the total runtime of the optimization versus 
$\ell(\alpha_*)$, where $\alpha_*$ is the solution of 
\eqref{exp 1 optimization problem} computed by the algorithms, or 
$\alpha_* = \alpha_{\rm rand}$. 
Runtime values associated with colored markers 
in Figure \ref{figure runtime} do not include
the computational time ($O(D^3)$) for computing 
the constraints matrix,  
which is needed to implement the nonnegativity- and triangle- 
inequalities in {\bf cvx}.
The total time is indicated, on the same plot, by black markers.
Different marker shapes in Figures \ref{figure runtime} 
refer to different choices of $D$.

\subsection{Experiment 2: feature space approximation (MNIST data)}
\label{section experiment 2}
\begin{figure}[h!]
\vskip 0.2in
\begin{center}
\centerline{\includegraphics[width=\columnwidth]{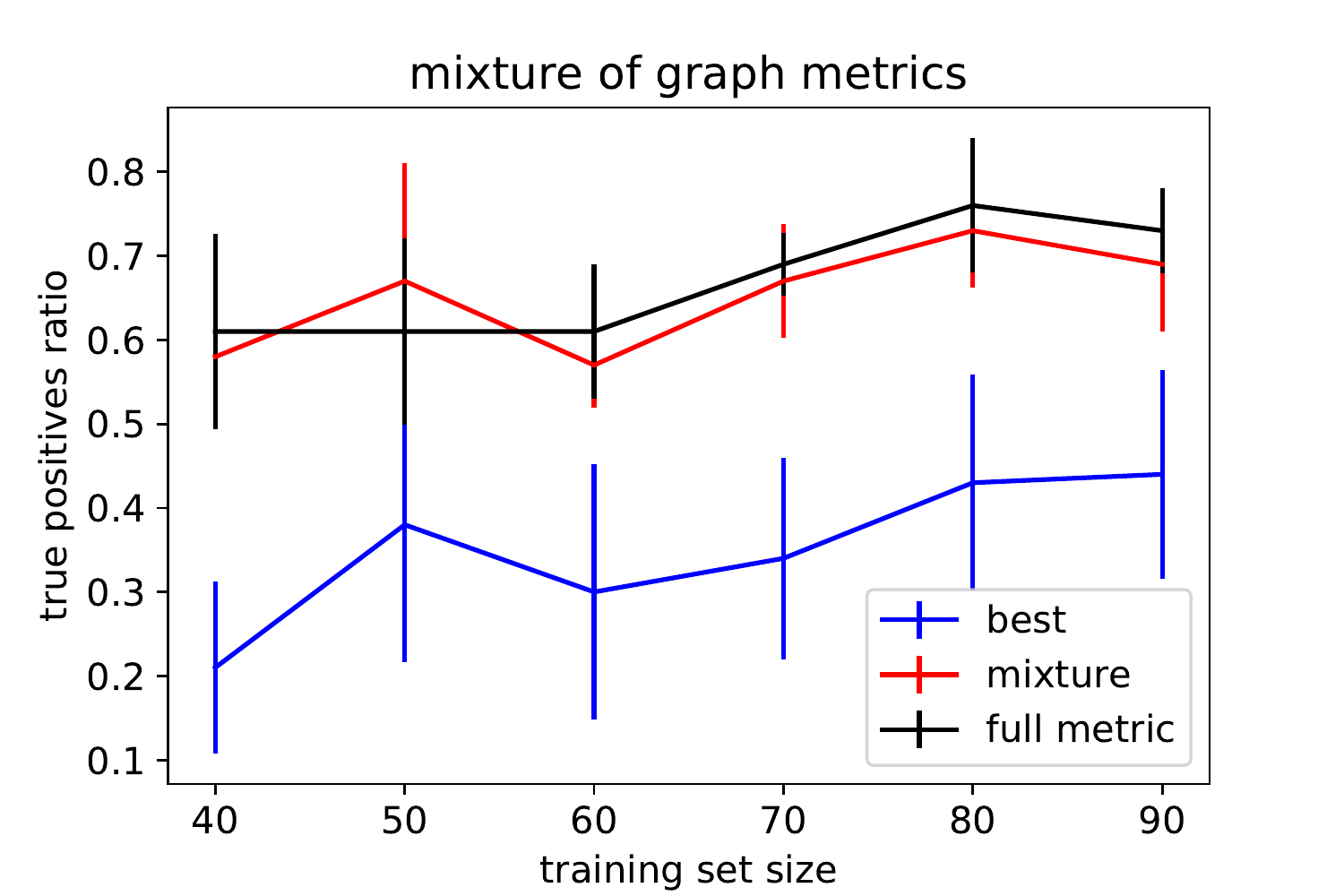}}
	\caption{Feature space approximation. 
	Predictive power of a $(k=1)$ Nearest Neighbor algorithm 
	based on a single graph metric ({\rm best}), an optimized 
	mixture of graph metrics ({\rm mixture})and the 
	full feature-space Euclidean metric ({\rm full}).
	Graph metrics are path-length distance metrics 
	associated with graphs ${\cal G}(A_\xi)$,  
	where the adjacency metrics $A_\xi$ 
	are defined by $[A_{\xi}]_{ij} = {\bf 1}{[M_{\rm true}]_ij < \xi}$, 
	for different choices of $\xi$. 
	The single graph metric is obtained by solving 
	\eqref{optimization algorithm 1} with simplex constraints
	on $\alpha$.
	Error bars represent standard deviations over 5 analogous 
	experiments.
	} 
\label{figure feature space approximation}
\end{center}
\vskip -0.2in
\end{figure}

Main contribution of the proposed method (see Section \ref{section motivations})
is the possibility of exploiting distinct non-Mahalanobis distance 
metrics simultaneously. 
In particular, this may allow one to combine the information coming from 
different graphs, each of them describing an independent aspects  
of a given system.
In social networks applications, for example, 
different graphs can be associated 
with different kinds of relationship between users.
The experiment shows how graph-related 
metrics can be merged to solve classification tasks.
For simplicity, we the classification is performed 
through a $k=1$-Nearest Neighbor algorithm, where 
predicted labels for test set instances 
are the labels of their closest neighbour in the train set.
We sample from the MNIST data base increasingly 
large train data sets, ${\cal D}^{(D)}_{\rm train}$ 
($D = |{\cal D}^{(D)}_{\rm train}| = 40, 50, 60, 70, 80, 90$), 
and a test data set ${\cal D}_{\rm test}$ of size 
$D_{\rm test} = 20$.
All data sets contain $D$ grey-scale images and the 
corresponding labels, i.e. 
${\cal D}_{\rm train}^{(D)} = \{ (x^{(i)}, y^{(i)} \}$ 
($i = 1, \dots, D$, idem ${\cal D}_{\rm test}$).
For each $D$, we use the images in ${\cal D}_{\rm train}^{(D)}$
to compute the corresponding feature-space Euclidean metric, 
$M_{\rm true} \in {\mathbf R}^{D \times D}$,
and a set of graph metrics 
$M_{r}$ ($r=1, \dots, R$), defined as 
in \eqref{exp 1 metrics} with $x^{(i)} \in {\cal D}^{(D)}_{\rm train}$ 
($i = 1, \dots, D$). 
We assume that $M_{\rm true}$ is not available and 
train a linear combination of the graph metrics, $M_\alpha$, defined as 
\eqref{exp 1 optimization problem}.
The possibly negative weight of the linear combination, $\alpha$,
are the solution the linear metric-constrained 
optimization problem \eqref{optimization algorithm 1}.
The solution is obtained through the algorithm 
sketched in Algorithm \ref{algorithm 1}.
Finally, we use ${\cal D}_{\rm test}$ and a $k=1$ Nearest Neighbor 
algorithm to compare the predictive 
performance of the trained mixture, $M_{\alpha_*}$ and 
two other metrics, $M_{\rm true}$ and $M_{\rm best}$ (see below).
More precisely, we let 
\begin{equation}
	y_{\rm predicted}^{(i')} = y^{(i_*)}, \quad 
	i_* = {\rm arg} \min_{i=1, \dots, D} \bar M^{(i')}_{(D + i')i},  
\end{equation}
where $i' = 1, \dots, D_{\rm test}$,  
$\bar M \in \{ \bar M^{(i')}_{\alpha_*}, 
\bar M^{(i')}_{\rm true}, 
\bar M^{(i')}_{\rm best} \}$,  
$\bar M_{u}^{(i')} \in {\mathbf R}^{(D + 1) \times (D + 1)}$ 
($i'=1, \dots, D$, $u = \alpha_*, {\rm true}, {\rm best}$) 
are defined as in \eqref{exp 1 metrics} on the augmented 
data set ${\cal D}_{\rm train}^{(D)} \cup \{ (x^{(i')}, y^{(i')}) \}$, 
$\alpha_*$ is the optimal solution of \eqref{optimization algorithm 1}
and $\bar M^{(i')}_{\rm best}= \bar M^{(i')}_{r_*}$,
with
$r_* = {\rm arg} \min_{r} \ell({\bf e}_{r})$. 
Figure \ref{figure feature space approximation} shows the 
performance of the three models versus the size of the 
training set.

\subsection{Experiment 3: semi-supervised learning (Citeseer data)}
\label{section experiment 3}
\begin{figure}[h!]
\vskip 0.2in
\begin{center}
\centerline{\includegraphics[width=\columnwidth]{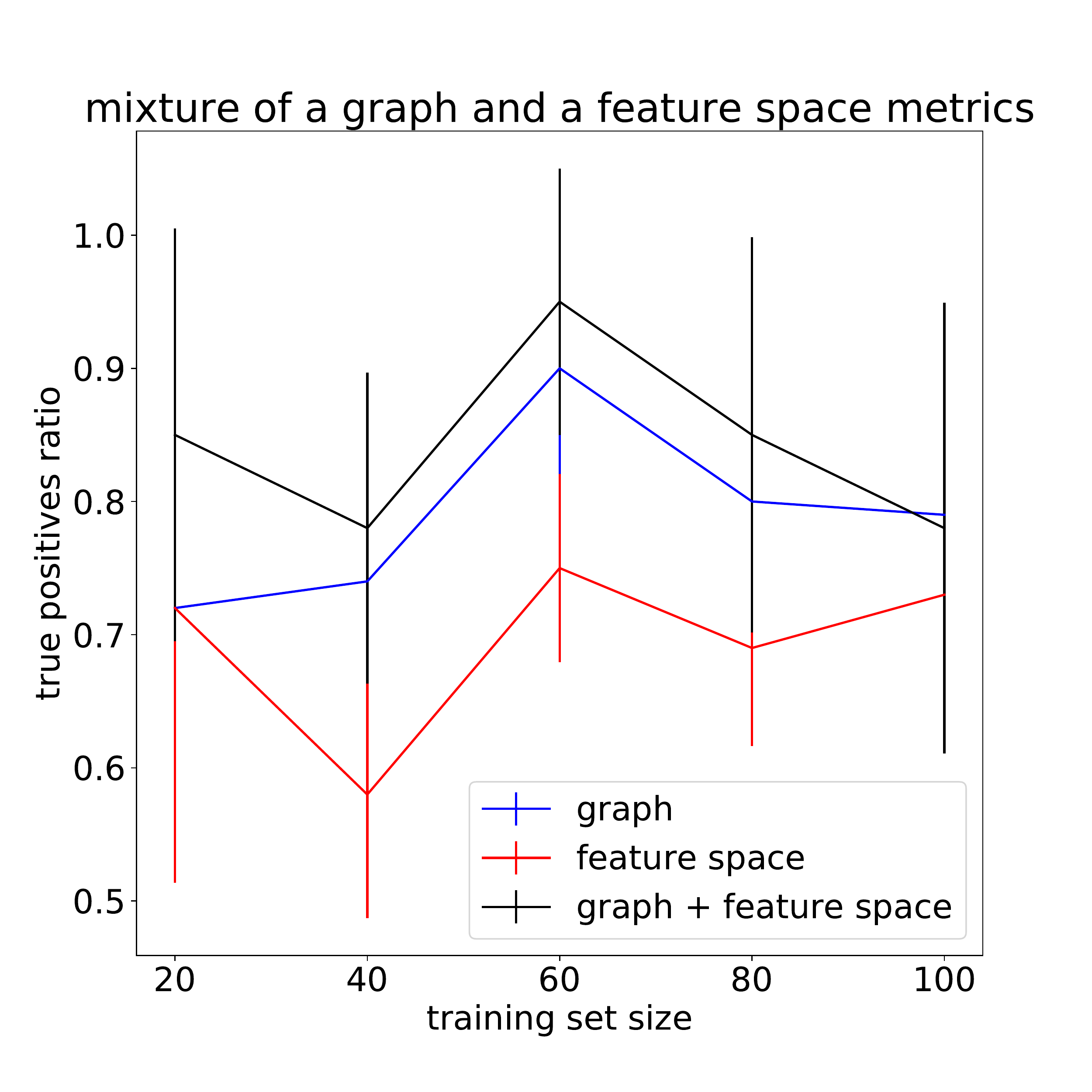}}
	\caption{Semi-supervised learning. 
	Predictive power of the $(k=1)$ Nearest Neighbor algorithm 
	based on a single graph metric, {\bf graph}, a single 
	(Mahalonobis) feature-space metric, {\bf feature space}, and 
	an optimized mixture of the two, {\bf graph + feature space}.
	The graph metric is a path-length distance metrics computed 
	on the citation sub-graph associated with a subset of the 
	Citeseer scientific papers.
	The feature-space metric is defined as the 
	Euclidean distance between 
	the corresponding (PC-projected) word-embedding vectors.
	Error bars represent standard deviations over 5 analogous 
	experiments.
	} 
\label{figure semi-supervised learning}
\end{center}
\vskip -0.2in
\end{figure}

Most common applications of metric learning from structured data 
are to networks analysis.
The problem is often cast as a semi-supervised learning problem, 
where the task is to predict 
a set of node labels, given their 
node attributes, a set of surrounding labelled nodes 
and the full edge structure
of the graph.
In those settings, it is natural to expect the prediction power 
of classification algorithms to be greatly improved 
if node and edge information can be effectively combined.
We test the proposed algorithm on such a semi-supervised learning
problem with $D_{\rm test} = 20$ missing 
labels and training sets of different sizes.
For both labelled and unlabelled nodes 
we have access to vectorized node attributes  
(the PC-projected word embedding of the Citeseer articles).
The complete node- and edge-structure of the graph (the citation network) 
is also available during training and testing.
In each experiment, we select sets of labelled 
articles from the Citeseer data base, 
${\cal D}^{(D + D_{\rm test})}_{\rm train}$,  
($D = 20, 40, 60, 80, 100$) and extract the associated citation 
sub-graphs ${\cal G}^{(D + D_{\rm test})}_{\rm train}$.
For all $D$, we use the citation sub-graphs, 
${\cal G}^{(D + D_{\rm test})}_{\rm train}$, 
and the node features, 
$x^{(i)} \in {\cal D}^{(D + D_{\rm test})}_{\rm train}$ 
to compute two training metrics: 
$M_{\rm graph} \in {\mathbf R}^{(D + D_{\rm test}) \times (D + D_{\rm test})}$, 
defined by 
$[M_{\rm graph}]_{ij} = {\rm dist}_{\rm path}(i, j, {\cal G}^{(D)})$ 
($i, j=1 \dots, D + D_{\rm test}$), 
and $M_{\rm feature}\in {\mathbf R}^{(D + D_{\rm test}) \times (D + D_{\rm test})}$, 
defined by $[M_{\rm feature}]_{ij} = \| x^{(i)} - x^{(j)} \|^2$.
($i, j=1 \dots, D + D_{\rm test}$).
We use the first $D$ (labelled) nodes to train a mixture model 
$M_{\alpha} = \alpha_{\rm graph} M_{\rm graph} + 
\alpha_{\rm feature} M_{\rm feature}$ and the remaining `unknown' 
labels for testing.
The predictive performance of a $(k = 1)$ Nearest Neighbor 
algorithm based on $M_{\alpha}$ is compared with 
the same algorithm based on $M_{\rm graph}$ and $M_{\rm feature}$. 
Training is performed as described in Section \ref{section experiment 2} 
by solving \eqref{optimization algorithm 1} 
through Algorithm \ref{algorithm 1}.
Figure \ref{figure semi-supervised learning} shows the 
performance of the three models versus the size of the 
training set.

\subsection{Results}
\paragraph{Experiment 1}
For small $D$, the performance of {\bf path} (the proposed method) 
is equivalent to the performance of a state-of-the-art 
quadratic-optimization solver, {\bf cvx}, 
in terms of accuracy and runtime.
This s remarkable, as {\bf path} is a general 
solver, i.e. it can be used for minimizing any (possibly non-convex)
metric-constrained function, while {\bf cvx} relies on the 
specific quadratic form of the objective.
The proposed method can also handle cases where 
a direct optimization fails (for memory problems 
when $D > 70$ on our machine).
For large $D$, the quality of the output produced 
by {\bf path} slightly increases, as it may be expected 
as the training set is larger, but computational 
times remain feasible.
Performance improvements could be obtained by implementing 
a reduced-size early stopping test, 
where $\ell(\alpha)$ defined in \eqref{exp 1 optimization problem} 
is replaced by  $\ell_{\rm test}(\alpha) = \sum_{(i,j) \in {\cal I}_{\rm test}} 
[M_{\rm true} - M_{\alpha}]^2_{ij}$, with ${\cal I}_{\rm test}$ 
being a small set of index pairs.
Tuning the learning parameters for each 
specific $D$ can also reduce the computational time and 
increase the output quality.

\paragraph{Experiment 2}
The reported performance of {\bf mixture} shows that a 
combination of graph metrics can be statistically 
equivalent to a full feature-space metric.
It is important to stress here that 
goal of our simulation was not to show that 
graph-based metrics can be better than feature-space metrics.
In our settings, this would make little sense as all 
graph-based metrics 
are obtained by cutting out some of the information 
contained in the full feature-space metric, $M_{\rm true}$.
The feature-space metric should  
be looked at as a `ground truth' model, 
which can be naturally expected to
achieve the best performance.
What we want to show, instead, is that the proposed algorithm 
is actually able to combine the information encoded 
by distinct graph structures in a consistent way.
This is underlined by the statistically robust 
performance gap between {\bf best} and {\bf mixture}. 

\paragraph{Experiment 3}
Results suggest that merging information from a vector-valued 
feature space and given graph structures is often profitable.
As in the previous experiments, we have considered minimal 
settings for clarity reasons. 
The overall predictive performance of the model 
would obviously be increased if more feature-space and 
graph-based metrics are added to the mixture.
For example, $M_{\rm feature}$ could be replaced 
by a set of pre-optimized Mahalanobis metrics 
(obtained through usual metric-learning methods 
on the same data set) and the path-length metric 
by a set of different graph-based dissimilarity matrices.
\footnote{Popular choices include the Jaccard index 
or Resource Allocation dissimilarity measures.}
As expected, the benefits of a combined approach
are more remarkable when the size of the training 
data set is small and meaningful independent 
information can 
be extracted from both $M_{\rm feature}$ and $M_{\rm graph}$.
The slight performance's drop of {\rm mixture} 
for increasing sizes of the training data set 
maybe due to i) a certain redundancy between the 
two metrics or ii) the fact that the algorithm 
has not completely converged.
Convergence problems for large values of $D$ 
probably arise because we 
run Algorithm \ref{algorithm 1} with a single 
set of optimization parameters 
for all values of $D$.

\section{Discussion}
The main contribution of this work can be summarized 
as follows.
We propose a new algorithm for optimizing a general 
objective under metric constraints and provide theoretical 
arguments for analyzing the algorithm's convergence in a 
specific but useful case.
We describe how the scheme can be 
used in network-based application for combining 
heterogeneous information 
extracted from the graph structure and 
the feature space associated with the node attributes. 
We run simple experiments to show that the predictive
power of simple classification algorithm is indeed boosted 
by merging a set of Mahalanobis and non-Mahalanobis metrics.

Possible future extensions of this work go in three directions.
\paragraph{Large scale approximation}
As the computational bottleneck of the algorithm is the 
computation of single-pair shortest paths, 
the overall speed of the algorithm may be increased 
by considering large-scale approximations of the metrics or 
a faster versions of the Dijkstra algorithm.
\paragraph{Mixture of local metrics} 
It has been shown that local metrics may outperform 
global ones in different tasks.
The proposed approach can be used to define data-dependent, 
i.e. local, linear combination of pre-learned local metrics.
This would be an interesting but challenging application, as the 
objective function would become nonlinear in the parameters. 
\paragraph{Nonlinear combination of metrics}
Since the proposed method does not require the linearity 
or convexity of the metric model 
possibly nonlinear and more flexible functions of the 
input matrices may be explored, 
e.g. by letting $M_{\alpha} = \phi_{\alpha}(\{ M_r \})$, where 
$\phi_\alpha$ is a metric-projected neural network.

\bibliographystyle{apalike}
\bibliography{refs}
\end{document}